\documentclass[journal]{IEEEtai}

\usepackage[colorlinks,urlcolor=blue,linkcolor=blue,citecolor=blue]{hyperref}

\usepackage{color,array}

\usepackage{graphicx}

\usepackage{microtype}
\usepackage{graphicx}
\usepackage{subfigure}
\usepackage{booktabs} 
\usepackage{amssymb}
\usepackage{amsfonts}
\usepackage{MnSymbol}
\usepackage{multirow}
\usepackage{amsmath}
\usepackage{bm}
\usepackage{color}
\usepackage{array}
\newcolumntype{C}[1]{>{\centering}p{#1}}
\usepackage{algorithm,algorithmic}

\usepackage{ntheorem}

\begin{document}

\title{Soft-Masked Semi-Dual Optimal Transport for Partial Domain Adaptation}

\author{Yi-Ming Zhai,~Chuan-Xian Ren$^*$,~Hong Yan,~\IEEEmembership{Fellow, IEEE}
\thanks{This work was supported in part by National Natural Science Foundation of China (Grant No. 62376291), in part by Guangdong Basic and Applied Basic Research Foundation (2023B1515020004), in part by Guangdong Province Key Laboratory of Computational Science at Sun Yat-sen University (2020B1212060032), in part by the Hong Kong Innovation and Technology Commission (InnoHK Project CIMDA), and in part by the Hong Kong Research Grants Council (Project 11204821).}
\thanks{Y.M. Zhai and C.X. Ren are with the School of Mathematics, Sun Yat-Sen University, Guangzhou, 510275, China. H. Yan is with the Department of Electrical Engineering, City University of Hong Kong, Hong Kong.
C.X. Ren is the corresponding author (email: rchuanx@mail.sysu.edu.cn).}}

\markboth{Journal of IEEE Transactions on XXX, Vol. 00, No. 0, Month 2025}
{First A. Author \MakeLowercase{\textit{et al.}}: Bare Demo of IEEEtai.cls for IEEE Journals of IEEE Transactions}

\maketitle

\begin{abstract}
Visual domain adaptation aims to learn discriminative and domain-invariant representation for an unlabeled target domain by leveraging knowledge from a labeled source domain. Partial domain adaptation (PDA) is a general and practical scenario in which the target label space is a subset of the source one. The challenges of PDA exist due to not only domain shift but also the non-identical label spaces of domains. In this paper, a Soft-masked Semi-dual Optimal Transport (SSOT) method is proposed to deal with the PDA problem. Specifically, the class weights of domains are estimated, and then a reweighed source domain is constructed, which is favorable in conducting class-conditional distribution matching with the target domain. A soft-masked transport distance matrix is constructed by category predictions, which will enhance the class-oriented representation ability of optimal transport in the shared feature space. To deal with large-scale optimal transport problems, the semi-dual formulation of the entropy-regularized Kantorovich problem is employed since it can be optimized by gradient-based algorithms. Further, a neural network is exploited to approximate the Kantorovich potential due to its strong fitting ability. This network parametrization also allows the generalization of the dual variable outside the supports of the input distribution. The SSOT model is built upon neural networks, which can be optimized alternately in an end-to-end manner. Extensive experiments are conducted on four benchmark datasets to demonstrate the effectiveness of SSOT.
\end{abstract}

\begin{IEEEImpStatement}
Domain adaptation is crucial in computer vision and pattern recognition. Specifically, Partial Domain Adaptation (PDA) is a practical scenario with non-identical label spaces across domains. However, prevailing adversarial learning-based PDA methods may suffer from training instability and mode collapse. Despite the wide application of the Optimal Transport (OT) algorithm in unsupervised domain adaptation, its effective extension to PDA remains a challenge. In this work, we propose an OT framework tailored explicitly for PDA, which effectively mitigates label shift and achieves a class-wise domain alignment in the shared feature space. Notably, the proposed network parameterized OT solver facilitates efficient handling of large-scale OT problems without imposing computational burdens. Extensive experimental evaluations against several SOTA methods demonstrate the superior performance and efficacy of our proposed methodology.
\end{IEEEImpStatement}


\IEEEpeerreviewmaketitle
\section{Introduction}
\label{sec:introduction}

\IEEEPARstart{S}{ufficient} labeled data are needed in training discriminative and robust models, which have wide applications in visual-based machine learning. However, the collection of annotated data is labor-intensive and time-consuming. Besides, labeled data for some tasks is extremely expensive or impossible due to privacy or other issues. Fortunately, the big data era can provide sufficient labeled training data with related scenarios. However, there may exist dataset shift between the labeled and unlabeled data due to exploratory factors of datasets,~\textit{e.g.}, style, background, and camera views~\cite{pan2010survey,jhuo2012robust,courty2017optimal}. Visual domain adaptation is an appealing strategy to reduce the dataset shift between the related source and target domains, which has been successfully applied in image classification~\cite{PGCD2023}, image segmentation~\cite{SCAN}, object detection~\cite{SAN2021}, and many other tasks.

Various domain adaptation methods attempt to reduce the domain discrepancy by matching statistic moments of domains~\cite{Long2015Learning,sun2016return,Long2017Deep}, employing adversarial learning~\cite{Ganin2017Domain,2018GTA,ADDA_extension}, manifold learning~\cite{luo2022DMP} or minimizing the Optimal Transport (OT) distance between domains~\cite{bhushan2018deepjdot,zhang2019optimal}. These methods mostly focus on Unsupervised Domain Adaptation (UDA), which assumes the source and target domains have identical label space. However, this can be an unrealistic assumption in real-world applications since the labels of the target domain are unknown.

\begin{figure}[!t]
    \centering
    \includegraphics[width=0.95\linewidth]{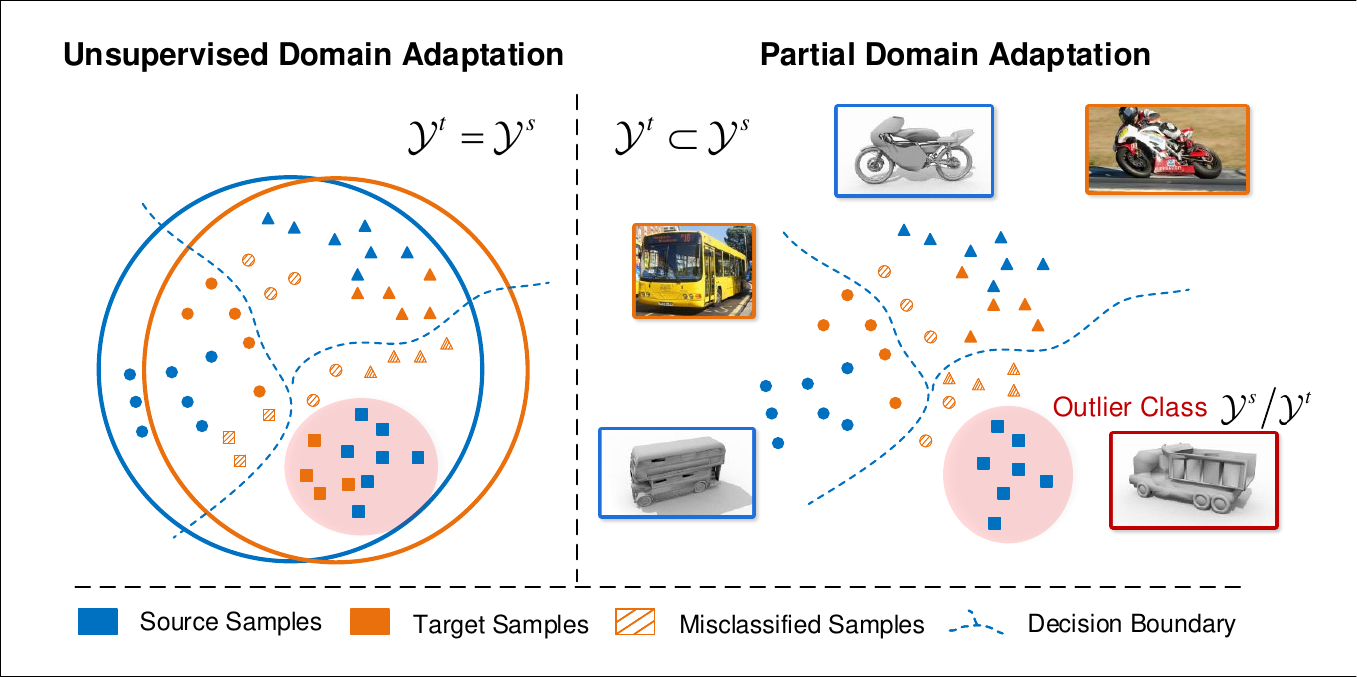}
    \caption{Illustration of the PDA problem. UDA assumes the source and target domains share the same label space, \textit{i.e.}, $\mathcal{Y}^t\!=\!\mathcal{Y}^s$. Direct application of the classifier learned on the source domain suffers from domain shift, as shown in the left. Compared with UDA, PDA assumes that the label space of the target domain is a subset of the source domain, \textit{i.e.}, $\mathcal{Y}^t\!\subset \!\mathcal{Y}^s$. The challenges of PDA are not only from domain shift but also the mismatch of the label spaces. Best viewed in color.}
    \label{fig:UDA_OT}
\end{figure}

Partial Domain Adaptation (PDA) is a more general and practical scenario, which assumes that the label space of the target domain is a subset of the source one, \textit{i.e.}, $\mathcal{Y}^t \subset \mathcal{Y}^s$. As shown in Fig.~\ref{fig:UDA_OT}, some categories in the source domain (\textit{e.g.}, truck) not belonging to the shared label space are referred as outlier classes. Besides, the outlier classes $\mathcal{Y}^s\backslash\mathcal{Y}^t$ are unknown. Thus, PDA is a scenario with an extreme label shift. For PDA, simply aligning the whole source and target domains may cause severe negative transfer since the target images may be misclassified to outlier classes. Therefore, the challenges of PDA not only come from dataset shifts but also the negative transfer due to the mismatch of label spaces.

To address these limitations, it is crucial for PDA methods to filter out the source outlier classes and improve the positive transfer across domains with shared label spaces. Most PDA methods are based on an adversarial learning framework. To be specific, methods~\cite{cao2018partial,cao2018partial_SAN,zhang2018importance} utilize class-wise domain discriminators or a source classifier to reweigh the source samples and learn domain-invariant features between the target and reweighed source domains. Besides, reinforcement learning~\cite{Chen_2020_CVPR,DARL2022} and similarity measurements~\cite{SPDA_2022} have also been exploited to reduce the importance of source outlier classes. These methods mainly achieve a global domain distribution alignment between the target and reweighed source domains while ignoring the discriminative structure of domains, which may lead to a misalignment between samples from different classes. Several methods have been proposed to seek a class-wise domain alignment, including uncovering intra- and inter-domain relationships via graph-based models~\cite{AGAN_2022} and matching clusters via reweighed maximum mean discrepancy (MMD)~\cite{TSCDA2021}, contrastive learning~\cite{CLCA_2023}, manifold learning~\cite{manifold2022} or co-training two diverse classifiers~\cite{CSDN_2023}. Though these methods can reweigh source samples, most of them do not mitigate the label shift between domains. With the conditional shift theory in~\cite{zhang2013domain}, the existence of label shift will lead to a bias of class-wise domain alignment in the latent feature space. Additionally, almost all PDA methods are based on adversarial learning, which may have problems of training instability and mode collapse. Besides, it is worth noting that many well-studied algorithms in DA have not been well extended to PDA, \textit{e.g.}, OT-based frameworks.

OT is a geometrically faithful metric for measuring the discrepancy between distributions~\cite{villani2008optimal}, which is also known as the Wasserstein distance or Earth Mover's distance. Compared with the Kullback-Leibler or Jensen-Shannon divergence, the Wasserstein distance incorporates the geometry information of the metric space via the cost function~\cite{arjovsky2017wasserstein}. Thus, it is appealing to learn discriminative and domain-invariant features by applying OT to domain adaptation. Specifically, various OT-based methods are mainly proposed for UDA, including matching domains by learning marginal invariant features~\cite{redko2017theoretical,courty2017optimal,zhang2019OT}, joint invariant features~\cite{courty2017joint,bhushan2018deepjdot}, and class-conditional invariant features~\cite{BuresNet}. Several OT-based PDA methods~\cite{JUMBOT_2021,mPOT_2022,MOT_2023} have been proposed, which are mostly based on Unbalanced OT (UOT)~\cite{chapel2021unbalanced}. Compared to classical OT, UOT relaxes the strict marginal constraints of transportation $\pi$, which allows it to achieve a partial matching in PDA. However, since the relaxation is applied to classical OT, the penalty may be unaffordable, and the relaxation will be inapplicable when the label shift is significant~\cite{MOT_2023}. Therefore, it is also meaningful and worthwhile to explore classical OT-based algorithms to deal with PDA.

In addition, existing domain adaptation methods based on OT are mainly constrained by two bottlenecks. First, the advantages of OT over other metrics rely on a high computational cost, \textit{e.g.}, the Sinkhorn algorithm has an $\mathcal{O}(n^2)$ complexity~\cite{cuturi2013sinkhorn}, which makes it not scale well to large-scale problems. Second, though OT is geometrically faithful in measuring distribution discrepancy, it does not take the label information into consideration. Besides, due to the mini-batch training manner of deep adaptation methods and mismatching label space in PDA, the sampled instances within mini-batches cannot fully reflect the real distribution. Then, the estimated optimal transport plan may be biased, and the negative transfer between domain may be more serious.

To tackle the above bottlenecks, in this paper, we propose a novel method named Soft-masked Semi-dual Optimal Transport (SSOT) for PDA by introducing a weighted semi-dual OT formulation and a soft mask mechanism. Instead of relying on adversarial learning, we explore the semi-dual formulation of the entropic regularized Kantorovich problem to make a domain distribution alignment in the shared label space. Specifically, we incorporate class-level importance weights into the semi-dual formulation to mitigate the significant label shift in PDA. Then, the corrected source domain is expected to share label distribution probabilities with the target domain. Further, a soft mask mechanism based on label predictions is proposed for reweighing the transport distance in OT, which attempts to map the images from the same class but different domains nearby in the shared feature space. To scale well on large-scale datasets, we employ a stochastic optimization algorithm for the semi-dual OT. Besides, we parameterize the optimization of OT with a neural network. Thus, the optimal Kantorovich potential, \textit{i.e.}, dual variable, can be explored more efficiently in a more compact parameter space. The whole framework of SSOT is built on deep learning, which leads to a mutual promotion between the optimization of OT and the learning of a more discriminative feature space. The main contributions of this paper are summarized as follows.

\begin{itemize}
\item[{1)}] A weighted semi-dual OT framework is proposed to mitigate the effect of significant label shift in PDA. By correcting the bias of label distributions across domains, the weighted semi-dual OT can learn a more accurate transportation between domains to promote a positive transfer.
\item[{2)}] To learn more discriminative features, we propose a soft-mask operation based on label information, and exploit the mask to reweigh the transport distances. The reweighed transport distance can reduce negative transfer by promoting a class-wise domain alignment.
\item[{3)}] By virtue of the powerful fitting ability of neural networks, it is expected to optimize OT with higher efficiency. The dual variable is re-parameterized by a two-layer fully connected network, instead of the vectored dual variable in the traditional semi-dual optimization problems.
\end{itemize}

The rest of this paper is organized as follows. In Section~\ref{sect:related-work}, we introduce related works of PDA and OT-based methods for domain adaptation. In Section~\ref{sect:method}, we provide the formulations of OT and details of SSOT for PDA. Extensive experiments and analysis are shown in Section~\ref{sect:experiment}. Finally, a conclusion is presented in Section~\ref{sect:conclusion}.

\section{Related Work}
\label{sect:related-work}

In this section, we briefly review the fruitful lines of PDA methods and the applications of OT on domain adaptation.

\textit{1) Partial Domain Adaptation}: To mitigate the negative transfer, existing PDA methods focus on filtering out the outlier classes and aligning domains with shared label spaces. In particular, adversarial learning has been employed by several researchers to deal with this problem. Cao~\textit{et al.}~\cite{cao2018partial} propose Partial Adversarial Domain Adaptation (PADA), which estimates the weights of source samples and incorporates the weights into the classifier and domain discriminator. Selective Adversarial Network~\cite{cao2018partial_SAN} trains separable domain discriminators for each class to achieve a fine-grained domain alignment. In Importance Weighted Adversarial Nets (IWAN), Zhang~\textit{et al.}~\cite{zhang2018importance} utilize an auxiliary domain classifier to identify the image similarities across domains. Chen~\textit{et al.}~\cite{DARL2022} propose a Domain Adversarial Reinforcement Learning (DARL) framework, which regards the source sample selection procedure as a Markov decision process and learns common feature space via domain adversarial learning. To further reduce the negative transfer brought by the class misalignment across domains, several methods~\cite{CLCA_2023,CSDN_2023} attempt to learn more discriminative features and achieve a class-wise domain alignment. Xu~\textit{et al.}~\cite{xu2019larger} propose a Stepwise Adaptive Feature Norm (SAFN) and demonstrate that task-specific features with larger norms are more transferable. Li~\textit{et al.}~\cite{li2021DRCN} propose a Deep Residual Correction Network (DRCN) to explicitly alleviate feature differences between domains and leverage a variant of MMD to reduce the discrepancy of each class across domains. Luo~\textit{et al.}~\cite{luo2022DMP} propose a Discriminative Manifold Propagation (DMP) method, which generalizes Fisher's discriminant criterion via the local manifold structures. Kim~\textit{et al.}~\cite{AGAN_2022} propose Adaptive Graph Adversarial Networks (AGAN) to exploit the relationships in intra- and inter-domain structures. Existing PDA methods mainly rely on adversarial learning to achieve domain alignment. The min-max training manner may have problems with training instability and mode collapse. Then, extending other frameworks for measuring the distribution discrepancy in PDA is still necessary.

\textit{2) Optimal Transport}: With solid theoretical guarantees~\cite{redko2017theoretical}, OT has been successfully applied to domain adaptation. The Kantorovich formulation of OT \cite{villani2008optimal} is commonly used in this context. Given two Polish probability spaces $(\mathcal{X},\mu)$ and $(\mathcal{Z},\nu)$, and two random variables $X\sim \mu$ and $Z \sim \nu$, the Kantorovich problem seeks the optimal transport plan $\pi$ to minimize the total transport cost between $\mu$ and $\nu$,
\begin{equation}
\label{eq:kantorovich_problem}
\underset{\pi}{\rm inf}\: \mathbb{E}_{(X, Z) \sim \pi}\left[ c(X,Z)\right ]~~~~{\rm s.t.}~~X \sim \mu,~Z \sim \nu, \notag
\end{equation}
where $\pi$ is a probability measure on $\mathcal{X}\times\mathcal{Z}$ with marginals $\mu$ and $\nu$, and $c(x, z):\mathcal{X}\times \mathcal{Z} \mapsto \mathbb{R}^+$ represents the cost of moving one unit mass from location $x$ to location $z$.

Further, Cuturi~\textit{et al.}~\cite{cuturi2013sinkhorn} derives a smoother version of the Kantorovich formulation by incorporating an entropy regularization term of the transport plan $\pi$, \textit{i.e.},
\begin{align}
\label{eq:regularized_Kantorovich}
\underset{\pi}{\rm inf}\: \mathbb{E}_{(X, Z) \sim \pi}\left[ c(X,Z)\right ]+\varepsilon R(\pi)~~~~{\rm s.t.}~~X \sim \mu,~Z \sim \nu,
\end{align}
where $R(\pi)=\mathbb{E}_{\pi\in\mathcal{X}\times\mathcal{Y}}[\ln(\frac{d\pi(x,y)}{d\mu(x)d\nu(x)})-1]$. The above formulation is linear and strictly convex, which can be solved efficiently with the Sinkhorn algorithm. However, the Sinkhorn algorithm still faces challenges in large-scale OT problems due to its $\mathcal{O}(n^2)$ complexity.

Various OT-based methods have been proposed for UDA. Courty \textit{et al.} \cite{courty2017optimal} introduced an optimal transformation between domains based on the Kantorovich formulation of OT. Zhang \textit{et al.} \cite{zhang2019OT} extended the Kantorovich problem to kernel space and applied the kernel Wasserstein distance for UDA with Gaussianity assumptions. JDOT \cite{courty2017joint} and DeepJDOT \cite{bhushan2018deepjdot} incorporated label information to reduce the discrepancy of joint feature/label distributions using OT. Xu \textit{et al.} \cite{xu2020reliable} incorporate spatial prototype information and intra-domain structures to construct a weighted Kantorovich formulation. Ren \textit{et al.} \cite{BuresNet} proposed a variant of OT distance to quantify the class-conditional distribution discrepancy between domains.

Recently, OT-based approaches have also been proposed for PDA. Gu~\textit{et al.} \cite{AR_2021} designed an adversarial reweighting model based on the Wasserstein distance to adjust the importance of the source domain. Fatras~\textit{et al.}~\cite{JUMBOT_2021} introduced a mini-batch strategy coupled with Unbalanced Optimal Transport (UOT) to mitigate the effect of outlier classes. Nguyen~\textit{et al.}~\cite{mPOT_2022} proposed partial OT for transportation between mini-batches to limit incorrect transportation. Luo \textit{et al.} \cite{MOT_2023} formulated a masked UOT approach for PDA, which characterizes label-conditioned sample correspondence and seeks class-wise domain alignment. UOT
and POT both rely on one regularized coefficient to penal
incorrect transportation with a lower cost. However, the penalty
may be unaffordable when cross-domain distributions are
extremely different.

In this paper, we propose a new method for PDA. SSOT exploits the semi-dual formulation of the entropy-regularized Kantorovich problem~\cite{genevay2016stochastic}, which is specifically designed for large-scale datasets. We also incorporate a soft mask and class-level importance weights based on label information into the semi-dual OT formulation to promote the exploration of the discriminative structure of domains and mitigate the severe label shift in PDA. Unlike the vector-based stochastic algorithms in \cite{genevay2016stochastic}, we use neural networks to parameterize the Kantorovich potentials in the semi-dual formulation, taking advantage of their strong fitting ability. Then, the whole framework is based on neural networks, where the OT metric optimization and domain distribution alignment can mutually promote each other.

\section{Methodology}
\label{sect:method}

In this section, we introduce the SSOT method. Section~\ref{subsec:weighted semi-dual OT} proposes a weighted semi-dual OT formulation for alleviating the label shift across domains. Section~\ref{subsec: mask} details the soft mask for distance reweighing and employs a network parameterization for the Kantorovich potential. The whole model and algorithm of SSOT are presented in Section~\ref{subsec:algorithm and optimization}.

In PDA, we assume that we have access to a labeled source domain $\mathcal{D}^{s}=\{\boldsymbol{x}_{i}^{s}, {y}_{i}^{s}\}_{i = 1}^{n_s}$ and an unlabeled target domain $\mathcal{D}^{t}=\{\boldsymbol{x}_j^t\}_{j=1}^{n_t}$, where $\boldsymbol{x}_i^s$, $\boldsymbol{x}_i^t$ represent images and ${y}_i^s\in \mathcal{Y}^s$ denotes the ground-truth label of $\boldsymbol{x}_i^s$. Specifically, we have $\mathcal{Y}^s=\{1,2,\dots, K\}$ and the label space of the target domain is a subset of the source domain, \textit{\textit{i.e.},} $\mathcal{Y}^t\subset\mathcal{Y}^s$. Besides, the source outlier classes $\mathcal{Y}^s\backslash\mathcal{Y}^t$ are unknown.

\begin{figure*}[!tb]
    \centering
    \includegraphics[width=0.95\linewidth]{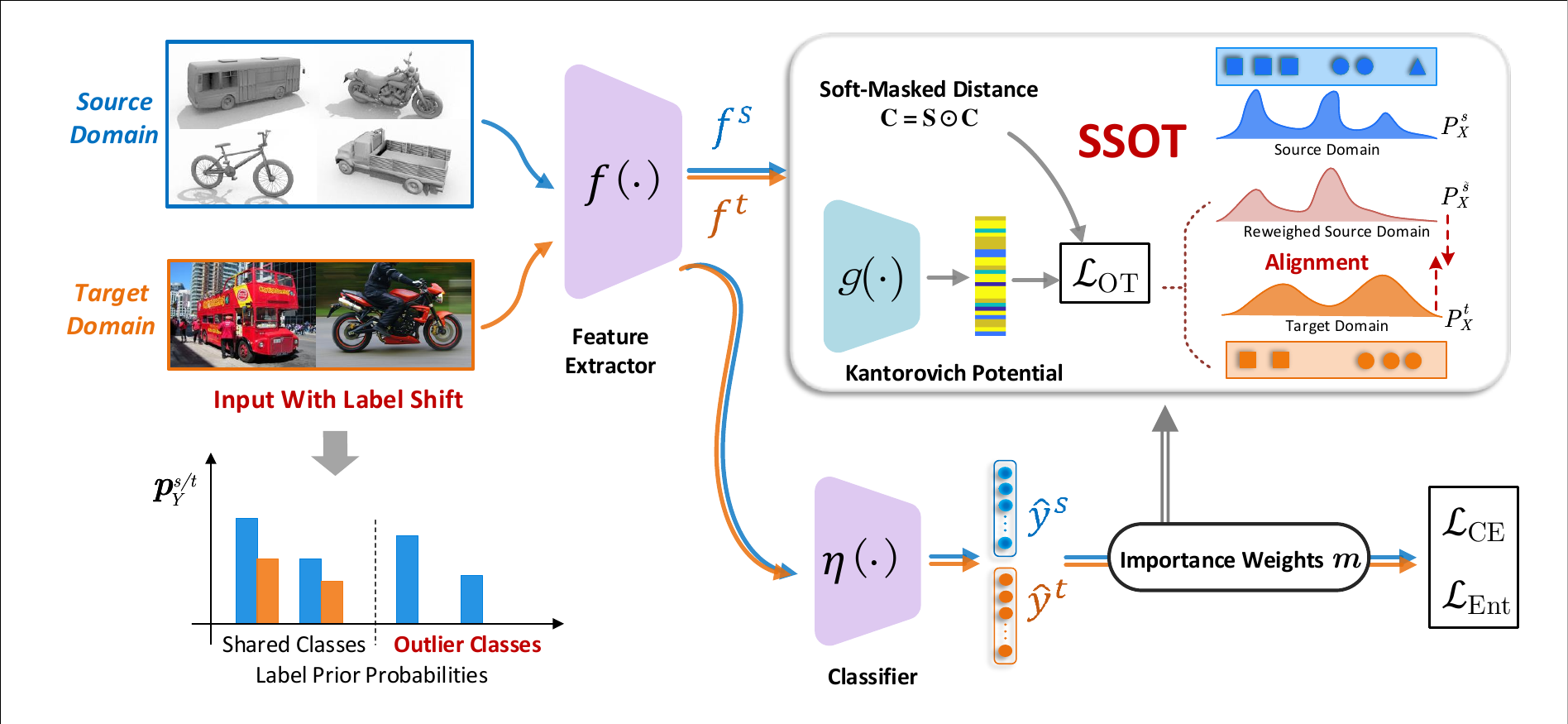}
    \caption{Flowchart of the proposed SSOT for PDA. The source and target domains share the network weights of the feature extractor $f(\cdot)$. To identify the source outlier classes and mitigate the label shift across domain, an importance weight $\boldsymbol{m}$ is introduced to reweigh the source domain. Then, the semi-dual OT formulation reduces the distribution discrepancy between the reweighed source domain $P_X^{\tilde s}$ and the target domain $P_X^t$. Besides, the cost matrix is enhanced by a soft-mask matrix to capture more class-relevant structures across domains. Specifically, the OT solver is parameterized by a neural network $g(\cdot)$ to approximate the Kantorovich potential. Best viewed in color.}
    \label{fig:network}
\end{figure*}

The entire network structure of SSOT consists of three parts, namely, the feature extractor network $f(\cdot)$, the classifier network $\eta(\cdot)$ and the Kantorovich potential network $g(\cdot)$. Their working flows are shown in Fig.~\ref{fig:network}. The feature extractor takes image $\boldsymbol{x}$ as input and outputs deep feature ${f(\boldsymbol{x})}$. The classifier maps the feature ${f(\boldsymbol{x})}$ as label prediction $\eta(f(\boldsymbol{x}))$. The parametrization network returns potential $g(f(\boldsymbol{x}))$ based on the deep feature. Since the model is updated in mini-batch, we suppose that there is a training batch $\mathcal{B}=\mathcal{B}^s\bigcup\mathcal{B}^t$, which contains a source batch $\mathcal{B}^{s}=\{\boldsymbol{x}_{i}^{s}, {y}_{i}^{s}\}_{i = 1}^{b_s}$ and a target batch $\mathcal{B}^{t}=\{\boldsymbol{x}_j^t\}_{j = 1}^{b_t}$. Here $b_{s/t}$ is the mini-batch size.

\subsection{Semi-Dual OT with Reweighed Label Distribution}
\label{subsec:weighted semi-dual OT}

Due to the existence of source outlier classes in PDA, directly aligning the marginal domain distributions, \textit{i.e.}, $P_X^s=P_X^t$, is easily prone to negative transfer between the source-outlier domain and target domain. Therefore, it is necessary to distinguish the source outlier classes and seek a domain alignment in the shared feature space.

Let $P_Y$ and $P_{X|Y}$ denote the label distribution and class-conditional distribution, respectively. Considering the label prior information, the domain distribution can be represented as mixtures of class-conditional distributions, \textit{i.e.},
\begin{equation*}
P_X^{s/t} = \sum\nolimits^{K}_{k=1}(\boldsymbol{p}_{Y}^{s/t})_k P_{X|Y=k}^{s/t},~~~\sum\nolimits^{K}_{k=1}(\boldsymbol{p}_{Y}^{s/t})_k = 1, \notag
\end{equation*}
where $\boldsymbol{p}_Y^{s/t}\in\mathbb{R}^K$ denotes the label prior probabilities (class weights) of the source/target samples, and its $k$-th element $(\boldsymbol{p}_{Y}^{s/t})_k$ represents the class weight of category $k$ in the domain. Due to the mismatched label distributions in PDA, \textit{i.e.}, $P_Y^s\neq P_Y^t$, it can be derived that $\boldsymbol{p}^s_Y \neq \boldsymbol{p}^t_Y$. Besides, the label space of the target domain is a subset of the source, \textit{i.e.}, $\mathcal{Y}^t\subset\mathcal{Y}^s$. Then, some elements of the target label prior probability $\boldsymbol{p}^t_Y$ are equal to zero.

To mitigate the effect of different label proportions across domains, it is reasonable to adjust the class weights of the source domain~\cite{zhang2013domain}. Taking into account a weight term $\boldsymbol{m}$, the adjusted source domain is supposed to have identical class weights with the target domain, which can be expressed as
\begin{align}
\label{eq:reweighed-source-domain}
P_X^{\tilde{s}} = \sum\nolimits^{K}_{k=1}\boldsymbol{m}_k (\boldsymbol{p}_{Y}^{s})_k P_{X|Y=k}^{s}.
\end{align}
Then, the importance weights $\boldsymbol{m}\in\mathbb{R}^{K}$ can be represented by the class ratios between source and target domains, \textit{i.e.},
\begin{align}
\label{eq:importance-weights-m}
\boldsymbol{m}_k=(\boldsymbol{p}_{Y}^t)_k/(\boldsymbol{p}_{Y}^s)_k,~~ \forall k\in\mathcal{Y}^s.
\end{align}
With the assistance of importance weights $\boldsymbol m$, the adjusted source domain and the target domain are suggested to have consistent label distributions, \textit{i.e.}, $P_Y^{\tilde s}= P_Y^t$.

Although OT has been widely used to deal with the UDA problem, it tends to match the feature distributions across domains via ${\rm OT}(P_X^s, P_X^t)$ and ignore the difference between label distributions of domains. However, PDA is a scenario with an extreme label distribution shift since the target label space is a subset of the source one. To reduce the negative transfer brought by the source outlier classes, SSOT is proposed to reduce the discrepancy between the reweighed source domain and target domain via optimizing ${\rm OT}(P_X^{\tilde s}, P_X^t)$ in the shared feature space.

The OT framework in SSOT is constructed on the entropy regularized Kantorovich formulation in~\eqref{eq:regularized_Kantorovich}. Though the Sinkhorn's iteration has achieved a lower computational cost, it still does not scale well to measures supported by a large number of samples. To tackle this bottleneck, we propose to apply a smoother semi-dual OT formulation~\cite{genevay2016stochastic} to PDA. By applying the Fenchel-Rockafellar's duality theorem on~\eqref{eq:regularized_Kantorovich}, a convex dual formulation is derived,
\begin{align}
\label{eq:dual-formulation}
\underset{u\in\mathcal{C}(\mathcal{X}),v\in\mathcal{C}(\mathcal{Z})}{\rm sup}\mathbb{E}_{X\sim\mu}[u(X)]+\mathbb{E}_{Z\sim\nu}[v(Z)]-F_ {\varepsilon}(u, v),
\end{align}
where $F_{\varepsilon}(u, v)=$
\begin{align}
& \left\{
\begin{aligned}
\nonumber
& \mathbb{I}_{U}(u,v),~&\varepsilon=0,\\
&\varepsilon\mathbb{E}_{(X,Z)\sim\mu\times\nu}\big[\exp\big(\frac{u(X)+v(Z)-c(X, Z)}{\varepsilon }\big)\big],~&\varepsilon > 0.
\end{aligned} \right.
\end{align}
Specifically, $\mathcal{C}(\mathcal{X})$ denotes the space of continuous functions on $\mathcal{X}$, and $\mathbb{I}_{U}(u,v)$ is an indicator function of the constraint set $U=\{(u,v);\forall(x,z)\in\mathcal{X}\times\mathcal{Z},u(x)+v(z)\leq c(x,z)\}$. Dual variables $u$ and $v$ are also known as Kantorovich potentials. When $\varepsilon = 0$, Eq.~\eqref{eq:dual-formulation} is the dual formulation of the Kantorovich problem. When $\varepsilon>0$, Eq.~\eqref{eq:dual-formulation} is the dual formulation of the regularized Kantorovich problem. In this case, $F_{\varepsilon}(u, v)$ is a smooth approximation of $\mathbb{I}_{U}(u,v)$.

The relation between $u$ and $v$ is obtained by applying the first order optimality condition of $v$ to~\eqref{eq:dual-formulation}, \textit{i.e.}, $u = v^{c,\varepsilon}$. Then, the semi-dual OT formulation can be derived by inserting the relational expression into~\eqref{eq:dual-formulation},
\begin{equation}
\label{eq:semi-dual-formulation}
\underset{v\in\mathcal{C}(\mathcal{Z})}{\rm sup}\mathbb{E}_{X\sim\mu}[v^{{c},\varepsilon}(X)]+\mathbb{E}_{Z\sim\nu}[v(Z)]-\varepsilon,
\end{equation}
where $\forall x\in\mathcal{X}$,
\begin{equation}
\label{eq:v-formulation}
v^{c,\varepsilon}(x)\!=\!
\left\{
\begin{aligned}
&\underset{z\in\mathcal{Z}}\min~{c}(x,z)-v(z),&\!\varepsilon=0,\\
&\!-\!\varepsilon\log\bigg(\mathbb{E}_{Z\sim\nu}\big[\exp\big(\frac{v(Z)-{c}(x, Z)}{\varepsilon }\big)\big]\bigg),&\!\varepsilon > 0.
\end{aligned} \right.
\end{equation}
When $\nu$ is a discrete distribution, the semi-dual formulation is a finite-dimensional concave maximization problem. Thus, it can be solved by gradient-based algorithms, which allow us to approximate the OT distance on large-scale datasets. Compared with the dual formulation, the semi-dual formulation is simpler since there is only one dual variable to be optimized. Therefore, the semi-dual OT formulation is employed in SSOT.

In SSOT, we map the samples into the latent feature space via the feature extractor $f(\cdot)$, and then employ the semi-dual OT formulation in~\eqref{eq:semi-dual-formulation} to measure the OT distance ${\rm OT}(P_X^{\tilde s}, P_X^t)$, \textit{i.e.},
\begin{equation*}
\underset{v\in\mathcal{C}(\mathcal{X})}{\rm sup}\:\mathbb{E}_{X^{s} \sim P_X^{\tilde s}} [v^{c,\varepsilon}(f(X^{s}))] + \mathbb{E}_{X^{t} \sim P_X^{\tilde t}}[v(f(X^{t}))] - \varepsilon,
\end{equation*}
where $v^{c,\varepsilon}(f(x^{s}))$ is similarly defined as~\eqref{eq:v-formulation}. With the importance weights $\boldsymbol{m}$, the above formulation aligns the feature distributions of the target domain and reweighed source domain in the shared feature space. Instead of exploring the optimal transport plan $\pi$, the semi-dual OT formulation seeks the optimal dual variable $v$, \textit{i.e.}, Kantorovich potential.

\subsection{Soft Mask Construction}
\label{subsec: mask}

Generally, the cost function $c(\cdot,\cdot)$ in OT is used to calculate the distance between samples from different distributions. Specifically, the OT distance incorporates the geometry information of the underlying support via the cost function~\cite{zhang2019optimal}. With the feature extractor $f(\cdot)$, the features of the source and target samples can be represented by
\begin{align*}
  \boldsymbol{f}_i^s = f(\boldsymbol{x}_i^s),~~\boldsymbol{f}_i^t = f(\boldsymbol{x}_i^t).
\end{align*}
Then, the cost matrix $\mathbf{C}\in\mathbb{R}^{b_s \times b_t}$ for each batch can be formulated by
\begin{align*}
  \mathbf{C}_{ij}= c(\boldsymbol{f}_i^s, \boldsymbol{f}_j^t),
\end{align*}
where the cost function $c(\cdot,\cdot)$ is usually specified as the squared Euclidean distance, \textit{i.e.}, $\mathbf{C}_{ij}=\|\boldsymbol{f}_i^s- \boldsymbol{f}_j^t\|^2_2$.

It is worth noting that the cost function directly considers the distance between samples across domains while ignoring the label information. Then, samples within mini-batch are difficult to fully reflect the real domain distributions, which may learn a biased data structure and lead to a misalignment between samples from the same class but different domains. Besides, seeking a class-wise domain alignment across domains is crucial for positive transfer. Overall, there is a strong motivation to define a label information-based mask on the cost matrix and promote the correct transportation between intra-class samples.

An intuitive idea is to split samples into class-wise clusters and construct multiple OT problems to reduce the discrepancy between clusters. Given samples $\{\boldsymbol{x}_i^s\}_{i=1}^{b_s}$ and $\{\boldsymbol{x}_j^t\}_{j=1}^{b_t}$, a hard mask matrix $\mathbf{H}\in\mathbb{R}^{b_s\times b_t}$ can be defined as
\begin{equation*}
\label{Eq:hard_matrix}
\mathbf{H}_{ij} = \left\{
\begin{array}{ccl}
1, &   & \mbox{if } y_i^s = y_j^t,\\
+\infty, &   & \mbox{else}.
\end{array}\right.
\end{equation*}
Then, we can obtain a masked cost matrix as $\tilde{\mathbf{C}}=\mathbf{C}\odot\mathbf{H}$, where $\odot$ represents the Hadamard product. With the mask $\mathbf{H}$, the transport cost for the inter-class sample pairs will be enlarged to infinity. Luo \textit{et al.}~\cite{MOT_2023} apply such a hard mask to UOT and theoretically derive that the masked UOT can seek a class-wise domain alignment. It is reasonable since the masked cost matrix $\tilde{\mathbf{C}}$ ensures that the optimal transport plan $\pi$ only assigns values for the intra-class sample pairs. However, the ground-truth labels of target samples $y_j^t$ are unavailable in PDA. In practice, pseudo labeling is an effective strategy in unsupervised learning~\cite{xie2018MSTN}. It is worth noting that the mask matrix $\mathbf{H}$ requires hard pseudo-labels for the target samples, which may be error-prone in the training process.

In this work, we propose to construct a soft mask matrix $\mathbf{S}\in\mathbb{R}^{b_s\times b_t}$ based on probability predictions,
\begin{equation}
\label{eq:score_m}
    \mathbf{S} = \mbox{softmax} \left[\mathbf{1}- ({\eta(\mathbf{F}^s)})^T  ({\eta(\mathbf{F}^t)}) \right],
\end{equation}
or
\begin{equation*}
    \mathbf{S}_{ij} = \frac{\exp{(1-\eta({{\boldsymbol{f}_i^s})}^T{\eta(\boldsymbol{f}_j^t}))}}{\sum_{j=1}^{b_t}\exp{(1-{\eta({\boldsymbol{f}_i^s})}^T{\eta(\boldsymbol{f}_j^t}))}},
\end{equation*}
where $\mathbf{F}^s=[\boldsymbol{f}_1^s,\boldsymbol{f}_2^s,\ldots,\boldsymbol{f}_{b_s}^s]^T\in\mathbb{R}^{b_s\times K}$, $\mathbf{F}^t\in\mathbb{R}^{b_t\times K}$ and $\mathbf{1}\in\mathbb{R}^{b_s\times b_t}$ is an all-ones matrix. The soft mask mechanism takes probability predictions $\eta(\mathbf{F}^s)$ and $\eta(\mathbf{F}^t)$ as inputs, and outputs a soft mask matrix $\mathbf{S}$, which is used as a mask operator. Then, the masked cost matrix can be formulated as
\begin{equation}
    \label{eq:attn_weighted}
    \tilde{\mathbf{C}}= \mathbf{S}\odot\mathbf{C}.
\end{equation}
Then, it deduces a reweighed distance metric $\tilde{c}(\boldsymbol{f}_i^s, \boldsymbol{f}_j^t)=\mathbf{S}_{ij}c(\boldsymbol{f}_i^s, \boldsymbol{f}_j^t)$, which is used to define a label information enhanced transport distance. The soft mask $\mathbf{S}$ makes a probabilistic adjustment to the cost matrix based on probability predictions, which can dynamically employ discriminative information to modify the data structure within each mini-batch. The adaptively adjusted cost matrix results in a transport plan that is expected to approximate the actual scenario. The parameter update of this mask mechanism is included in the adaptive process.

With the masked distance metric $\tilde{c}(\cdot,\cdot)$ defined in~\eqref{eq:attn_weighted}, the semi-dual OT formulation in SSOT can be rewritten as \footnote{For simplicity, the features of the source and target domains are abbreviated as ${f}^s$ and ${f}^t$, respectively.},
\begin{equation}
\label{eq:masked-semi-dual-expectation}
\underset{v\in\mathcal{C}(\mathcal{X})}{\rm sup}\:\mathbb{E}_{f^{s} \sim P_X^{\tilde s}} [v^{\tilde{c},\varepsilon}(f^{s})] + \mathbb{E}_{f^{t} \sim P_X^{\tilde t}}[v(f^{t})] - \varepsilon,
\end{equation}
where $v^{\tilde{c},\varepsilon}(f^{s})$ is also similarly defined as~\eqref{eq:v-formulation}. By optimizing the soft-masked OT distance between the target and reweighed source domain, our SSOT can learn both domain-invariant and class-discriminative features.

The masked semi-dual OT formulation in~\eqref{eq:masked-semi-dual-expectation} is an unconstrained concave maximization problem, which can be solved by stochastic gradient methods. In PDA, $P_X^s$ and $P_X^t$ are discrete distributions since they are only accessible through discrete samples. Thus, the dual variable $v$ can be initialized by a random vector (dimension equals the distributed sample size) and updated iteratively. Inspired by the strong fitting ability of neural networks, we propose to parameterize the dual variable $v$ with a neural network $g(\cdot)$. Denote the parameter of $g(\cdot)$ as $\mathbf{W}_g$. Based on the learned deep features, the masked semi-dual OT formulation between $P_X^{\tilde s}$ and $P_X^t$ in~\eqref{eq:masked-semi-dual-expectation} can be optimized w.r.t parameter $\mathbf{W}_g$,
\begin{equation}
\label{eq:semi-dual-expectation}
\underset{\mathbf{W}_g}{\rm sup}\:\mathbb{E}_{f^{s} \sim P_X^s} [g^{\tilde{c},\varepsilon}(f^{s};\mathbf{W}_g)] + \mathbb{E}_{f^{t} \sim P_X^t}[g(f^{t};\mathbf{W_g})] - \varepsilon,
\end{equation}
where $g^{\tilde{c},\varepsilon}(f^{s};\mathbf{W}_g)= $
\begin{align}
\left\{
\begin{aligned}
 \nonumber
&\mathop{\min}\limits_{f^{t}\in\mathcal{D}^t} \tilde{c}(f^{s},f^{t})-g(f^{t};\mathbf{W}_g),&\varepsilon = 0, \\
&-\varepsilon \log\bigg(\mathbb{E}_{f^{t} \sim \nu}\big[\exp{ \Big(\frac{g(f^{t};\mathbf{W}_g)-\tilde{c}(f^{s},f^{t})}{\varepsilon }\Big)}\big]\bigg),&\varepsilon > 0. \\
\end{aligned} \right.
\end{align}
Such a network parameterization allows an efficient and accurate approximation of the Kantorovich potential.

To present the calculation clearly, we reformulate~\eqref{eq:semi-dual-expectation} as a finite-dimensional optimization problem below.
Suppose that all the samples are uniformly sampled from corresponding probability simplex. Then, the empirical distributions of the source and target features are
\begin{align*}
P_X^s =\frac{1}{n_s}\sum_{i=1}^{n_s}\delta(\boldsymbol{f}_i^s),~~P_X^t=\frac{1}{n_t}\sum_{j=1}^{n_t}\delta(\boldsymbol{f}_j^t),
\end{align*}
where $\delta(\cdot)$ is the Dirac function. The empirical distribution of the reweighed source domain is adjusted as non-uniform $P_X^{\tilde s}$ by importance weights $\boldsymbol m$, \textit{i.e.},
\begin{align*}
P_X^{\tilde s}=\frac{1}{n_s}\sum_{i=1}^{n_s}\boldsymbol{m}_{y_i^s}\delta(\boldsymbol{f}_i^s),~~~\frac{1}{n_s}\sum_{i=1}^{n_s}\boldsymbol{m}_{y_i^s}=1,
\end{align*}
where $y_i^s$ is the label of feature $\boldsymbol{f}_i^s$ (sample $\boldsymbol{x}_i^s$). (The estimation of $\boldsymbol{m}$ is described in Section~\ref{subsec:algorithm and optimization}.) Then, with the masked cost matrix $\tilde{\mathbf{C}}$, the expectation maximization in~\eqref{eq:semi-dual-expectation} can be written as finite-dimensional optimization formulation, \textit{i.e.},
\begin{equation}
\label{eq:semi-dual-finite}
\underset{\mathbf{W}_g}{\rm max}~\mathcal{H}_{\varepsilon}(\mathbf{W}_g)=\frac{1}{n_s}\sum_{i=1}^{n_s} \boldsymbol{m}_{y_i^s} g^{\tilde{c},\varepsilon}(\boldsymbol{f}_i^s) + \frac{1}{n_t}\sum_{j=1}^{n_t}g(\boldsymbol{f}_j^t)-\varepsilon,
\end{equation}
where $g^{\tilde{c},\varepsilon}(\boldsymbol{f}_i^s)=$
\begin{align}
\left\{
\begin{aligned}
&\mathop{\min}\limits_{\boldsymbol{f}_j^t \in \mathcal{D}^T}\tilde{c}(\boldsymbol{f}_i^s,\boldsymbol{f}_j^t)-g(\boldsymbol{f}_j^t), &\varepsilon = 0, \notag\\
&-\varepsilon\log\bigg(\frac{1}{n_t}\sum_{j=1}^{n_t}\exp{ \Big(\frac{g(\boldsymbol{f}_j^t)-\tilde{c}(\boldsymbol{f}_i^s,\boldsymbol{f}_j^t)}{\varepsilon }\Big)}\bigg), &\varepsilon > 0.  \notag
\end{aligned} \right.
\end{align}
The maximization of $\mathcal{H}_{\varepsilon}(\mathbf{W}_g)$ is an unconstraint concave problem. Thus, we use stochastic gradient descent (SGD) to train the Kantorovich potential network $g(\cdot)$ by sampling batches $\mathcal{B}^s$ and $\mathcal{B}^t$. Since it is necessary to compute the gradient of $g^{\tilde{c},\varepsilon}(\cdot)$, the complexity of each iteration is $\mathcal{O}(b)$, where $b=\max(b_s, b_t)$.

In this way, the optimization of dual variables based on vector approaches changes to a training process of Kantorovich potential network $g(\cdot)$. Such an OT solver with network parameterization is consistent with the adaptive process based on deep learning, which provides a totally deep OT framework for domain adaptation. Besides, the network parametrization can simplify the algorithm calculation since the whole algorithm is trained by SGD in a mini-batch manner.

\subsection{Model and Numberical Optimization}
\label{subsec:algorithm and optimization}

In the PDA scenario, we aim to learn the feature extractor $f(\cdot)$ and classifier $g(\cdot)$ that can minimize the empirical risk on the target domain, \textit{i.e.},
\begin{equation*}
\varepsilon_t = \mathbb{E}_{(X^t,Y^t)}[\ell(x,y;f,\eta)],
\end{equation*}
where $\ell(\cdot)$ is the loss function (\textit{e.g.}, cross-entropy). However, the target domain is unlabeled. Since the support of the target domain is contained by that of the source domain, we can reformulate the target risk $\varepsilon_t$ as
\begin{align*}
\varepsilon_t &= \iint \ell(x,y;f,\eta)\frac{p_{x|y}^t p_y^t}{p_{x|y}^s p_y^s}p_{xy}^s{\rm d}x{\rm d}y\\
&=\mathbb{E}_{{(X^s,Y^s)}}[w(x,y)\ell(x,y;f,\eta)],
\end{align*}
where the weight $w(x,y)=\frac{p_{x|y}^t p_y^t}{p_{x|y}^s p_y^s}$. Then, the target risk can be changed to a weighted risk on the source domain. Specifically, the OT loss in SSOT seeks optimal transportation between intra-class samples, which promote a class-conditional distribution alignment, \textit{i.e.}, $P_{X|Y}^s=P_{X|Y}^t$. Thus, the weight can be approximately simplified as $w(y)=\frac{p_y^t}{p_y^s}$. It is worth noting that $w(k)=\boldsymbol{m}_k$, where the importance weight $\boldsymbol{m}$ is also necessary for constructing the reweighed source domain defined in~\eqref{eq:reweighed-source-domain}.

According to~\eqref{eq:importance-weights-m}, the estimation of label probabilities of domains is required for the estimation of the importance weights $\boldsymbol{m}$.
Since the source domain is labeled, the source label probability $\boldsymbol{p}_Y^s$ can be estimated by
\begin{align}
(\hat{\boldsymbol{p}}_Y^s)_k = \sum_{i=1}^{n_s}\frac{\mathbb{I}{y_i^s=k}}{n^s}, \notag
\end{align}
where $\mathbb{I}$ is an indicator function. However, $\boldsymbol{p}_Y^t$ cannot be estimated similarly since the target domain is unlabeled. In this paper, we estimate $\boldsymbol{p}_Y^t$ by averaging target predictions,
\begin{align}
\label{eq:class-weights-target}
\hat{\boldsymbol p}_{Y}^t=\frac{1}{n_t}\sum_{i=1}^{n_t}\eta(f(\boldsymbol{x}_i^t)),
\end{align}
where class weights $(\hat{\boldsymbol p}_{Y}^t)_k$ of these shared classes $\mathcal{Y}^s\bigcap\mathcal{Y}^t$ tend to be larger than these outlier classes $\mathcal{Y}^s\backslash\mathcal{Y}^t$. If class weights $\hat{\boldsymbol p}_{Y}^t$ is an optimal estimation, $(\hat{\boldsymbol p}_{Y}^t)_k$ of outlier classes $\mathcal{Y}^s\backslash\mathcal{Y}^t$ are supposed to be 0.
Overall, the importance weights $\boldsymbol m$ can be estimated by
\begin{equation}
\label{eq:importance weight estimation}
\hat{\boldsymbol m}_k=(\hat{\boldsymbol{p}}^t_Y)_k/(\hat{\boldsymbol{p}}^s_Y)_k.
\end{equation}

Then, given the importance weight $\hat{\boldsymbol{m}}$, the empirical weighted source risk associate with cross-entropy function $l_{ce}(\cdot,\cdot)$ can be expressed as
\begin{align}
\label{eq:pda-ce-loss}
\mathcal{L}_{\rm CE}(\mathbf{W})&\triangleq \frac{1}{n_s} \sum_{i=1}^{n_s} \hat{\boldsymbol m}_{y_i^s} l_{ce}(\eta(f(\boldsymbol{x}_i^s)), y_i^s) \notag \\
&= - \frac{1}{n_s} \sum_{i=1}^{n_s} \sum_{k=1}^{K} \hat{\boldsymbol m}_{y_i^s} {\boldsymbol y}^s_{ik}\log{\hat{\boldsymbol y}^s_{ik}},
\end{align}
where $\hat{\boldsymbol y}^s_i\!=\!\eta(f(\boldsymbol{x}_i^s))$ and $\boldsymbol{y}_i^s$ is a one-hot vector of $y_i^s$. Besides, $\hat{\boldsymbol y}^s_{ik}$ is the $k$-th element of $\hat{\boldsymbol y}^s_i$, which represent the prediction probability of source sample $\boldsymbol{x}_i^s$ belonging to the $k$-th class. The notation $\mathbf{W}$ denotes the parameters of the feature extractor $f(\cdot)$ and the classifier $\eta(\cdot)$. From~\eqref{eq:pda-ce-loss}, it can be seen that minimizing the weighted cross-entropy loss $\mathcal{L}_{\rm CE}$ can down-weight the contributions of outlier source samples to the classifier.

The entropy criterion is exploited to explore the intrinsic structure of the target domain. Mathematically, the target entropy loss denoted by $\mathcal{L}_{\rm Ent}$ is formulated as
\begin{eqnarray}
\label{eq:loss-ent}
\mathcal{L}_{\rm Ent}(\mathbf{W})\triangleq -\frac{1}{n_t} \sum^{n_t}_{j=1} \sum_{k=1}^{K} \hat{\boldsymbol y}^t_{jk}\log{\hat{\boldsymbol y}^t_{jk}},
\end{eqnarray}
where $\hat{\boldsymbol y}^t_{jk}$ is the prediction probability of target sample $\boldsymbol{x}_j^t$ belonging to the $k$-th class, where $\sum^{K}_{k=1}\hat{\boldsymbol{y}}_{jk}^s=1$ and $K=|\mathcal{Y}^{s}|$ is the number of classes.

With the predictions, the cost matrix can also be reweighed by the soft mask mechanism in~\eqref{eq:score_m}. Then, the Kantorovich network $g(\cdot)$ can be trained to approximate the general function via~\eqref{eq:semi-dual-finite}. With the learned parameters $\mathbf{W}_g$ of the Kantorovich network $g(\cdot)$, the optimal transport distance $W_{\varepsilon}({P_X^{\tilde s}},P_X^t)$ can be calculated by
\begin{eqnarray}
\nonumber
W_{\varepsilon}({P_X^{\tilde s}}\!,\!P_X^t)\!=\!\frac{1}{n_s}\sum_{i=1}^{n_s}\boldsymbol{\hat m}_{y_i^s}g^{\tilde{c},\varepsilon}(\boldsymbol{f}_i^s;\mathbf{W}_g) + \frac{1}{n_t}\sum_{j=1}^{n_t}g(\boldsymbol{f}_j^t;\mathbf{W}_g)-\varepsilon,
\end{eqnarray}
where $g^{\tilde{c},\varepsilon}(\cdot)$ is defined in~\eqref{eq:semi-dual-finite}. Then, the domain distribution discrepancy (\textit{i.e.}, the alignment loss) between the reweighed source domain and the target domain  can be measured by
\begin{equation}
\label{eq:loss-adapt}
    \mathcal{L}_{\rm OT}(\mathbf{W})\triangleq W_{\varepsilon}({P_X^{\tilde s}},P_X^t).
\end{equation}
By minimizing loss $\mathcal{L}_{\rm OT}$ w.r.t $\mathbf{W}$, it is expected to learn both transferable and discriminative features across domains under the guidance of weighted semi-dual OT.

Combining the above losses, the overall objective function of SSOT consists of three parts, namely the source classification loss $ \mathcal{L}_{\rm CE}$ in~\eqref{eq:pda-ce-loss}, domain adaptation loss $\mathcal{L}_{\rm OT}$ in~\eqref{eq:loss-adapt} and target entropy loss $\mathcal{L}_{\rm Ent}$ in~\eqref{eq:loss-ent}, which can be written as
\begin{align*}
\mathcal{L}_{\rm SSOT}(\mathbf{W}) = \mathcal{L}_{\rm CE}(\mathbf{W})+ \lambda_{\rm OT}\mathcal{L}_{\rm OT}(\mathbf{W})+ \lambda_{\rm Ent}\mathcal{L}_{\rm Ent}(\mathbf{W}),
\end{align*}
where $\lambda_{\rm OT}$ and $\lambda_{\rm Ent} > 0$ are trade-off hyper-parameters for balancing the effects of the three losses. The model reduces domain discrepancy by minimizing the optimal transport loss $\mathcal{L}_{\rm OT}$, and learns a discriminant classifier by minimizing the cross-entropy loss $ \mathcal {L}_{\rm CE}$. Further, the target entropy loss $\mathcal{L}_{\rm Ent}$ helps the model to explore a more discriminative feature space. With the importance weights, it is expected to filter out the source outlier classes and achieve domain alignment between features in the shared label space. The parameters of the feature extractor $f(\cdot)$ and classifier $\eta(\cdot)$, \textit{i.e.}, $\mathbf{W}$, will be learned by minimizing $\mathcal{L}_{\rm SSOT}(\mathbf{W})$ with SGD in a mini-batch manner.

The overall pipeline of SSOT for PDA is summarized in Algorithm~\ref{alg:SSOT}. Note that there are two loops in the algorithm, where the optimal transport module (w.r.t. $\mathbf{W}_g$) is a built-in loop. We update the parameters of the adaptive model and the Kantorovich potential parametrization, \textit{i.e.}, $\mathbf{W}$ and $\mathbf{W}_g$, in an alternative manner. To be specific, we fix the network parameters $\mathbf{W}$ and determine the optimal dual variable, and then fix the Kantorovich network parameters $\mathbf{W}_g$ to update the network parameters.

\begin{algorithm}[!t]
\caption{SSOT for PDA}
\label{alg:SSOT}
\begin{algorithmic}[1]
\REQUIRE{Source domain $\mathcal{D}^s\!=\!\{\boldsymbol{x}^s_i,{y}^s_i\}_{i = 1}^{n_s}$, target domain $\mathcal{D}^t\!=\!\{\boldsymbol{x}^t_j\}_{j=1}^{n_t}$, batch sizes $b_s$, $b_t$, OT weight $\lambda_{\rm OT}$, entropy weight $\lambda_{\rm Ent}$, learning rate $\alpha$, and regularized weight $\varepsilon$.}\\
\ENSURE{Networks parameters $\mathbf{W}_g$, $\mathbf{W}$, predictions $\{\hat{y}^t_j\}_{j = 1}^{n_t}$.}\\
\STATE Pre-train networks $f(\cdot)$ and $\eta(\cdot)$ via cross-entropy loss on the source domain $\mathcal{D}^s$;
\FOR{\textit{Adaptation iterations}}
\STATE Sample data from $\mathcal{D}^s$ and $\mathcal{D}^t$\\
~~~~~$\mathcal{B}^s=\{\boldsymbol{x}^s_i,{y}^s_i\}_{i=1}^{b_s}$,~~ $\mathcal{B}^t=\{\boldsymbol{x}^t_j\}_{j=1}^{b_t}$;\\
\STATE Forward propagate data\\
~~~~~$ \boldsymbol{f}=f(\boldsymbol{x})$,~~$\hat{\boldsymbol{y}}= \eta(\boldsymbol{f})$;\\
\STATE Estimate the soft mask matrix $\mathbf{S}$ via~\eqref{eq:score_m};\\
\STATE Reweigh the cost matrix $\mathbf{C}$ as~\eqref{eq:attn_weighted}\\
~~~~~$\tilde{\mathbf{C}}\leftarrow\mathbf{S}\odot\mathbf{C}$;\\
\STATE Forward propagate entire $\mathcal{D}^t$ without gradients; then, estimate $\hat{\boldsymbol{p}}^t_Y$ via~\eqref{eq:class-weights-target} and $\hat{\boldsymbol{m}}$ via~\eqref{eq:importance weight estimation};\\
\% {Fix $f(\cdot)$ and $\eta(\cdot)$, update $g(\cdot)$ for OT}
    \STATE Estimate the semi-dual formulation via~\eqref{eq:semi-dual-finite};
    \STATE Update: $\mathbf{W}_g \leftarrow \mathbf{W}_g + \alpha \nabla \mathcal{H}_{\varepsilon}(\mathbf{W}_g )$;\\
\% {Fix $g(\cdot)$, update $f(\cdot)$ and $\eta(\cdot)$ for adaptation}
\STATE Estimate the OT distance $\mathcal{L}_{\rm OT}$ via~\eqref{eq:loss-adapt};\\
\STATE Estimate the entropy-based loss\\
~~~~~$\mathcal{L}_{\rm CE}$ via~\eqref{eq:pda-ce-loss},~~$\mathcal{L}_{\rm Ent}$ via~\eqref{eq:loss-ent};\\
Compute the overall objective\\
~~~~~$\mathcal{L}_{\rm SSOT} =  \mathcal{L}_{\rm CE} + \lambda_{\rm OT} \mathcal{L}_{\rm OT} + \lambda_{\rm Ent} \mathcal{L}_{\rm Ent}$;\\
\STATE Update: $\mathbf{W}\leftarrow \mathbf{W}- \alpha \nabla \mathcal{L}_{\rm SSOT}(\mathbf{W})$.
\ENDFOR
\end{algorithmic}
\end{algorithm}

\section{Experiment Results and Analysis}
\label{sect:experiment}

\begin{table*}[!tb]
    \centering
    \caption{Accuracies (\%) on Office-Home and VisDA-2017 for PDA (ResNet-50).}
    \renewcommand{\tabcolsep}{0.235pc}
    \label{tab:office-home-visda-acc}
    \begin{tabular}{l|ccccccccccccc|ccc}
         \toprule
         \multirow{2}{*}{Method} & \multicolumn{13}{c|}{\textbf{Office-Home}} & \multicolumn{3}{c}{\textbf{VisDA-2017}}  \\
                                 & Ar$\rightarrow$Cl& Ar$\rightarrow$Pr& Ar$\rightarrow$Rw& Cl$\rightarrow$Ar& Cl$\rightarrow$Pr& Cl$\rightarrow$Rw& Pr$\rightarrow$Ar& Pr$\rightarrow$Cl& Pr$\rightarrow$Rw& Rw$\rightarrow$Ar& Rw$\rightarrow$Cl& Rw$\rightarrow$Pr& Mean &R$\rightarrow$S6 & S$\rightarrow$R6 & Mean\\
         \hline
         Source~\cite{He2016Deep}            & 46.3 & 67.5 & 75.9 & 59.1 & 59.9 & 62.7 & 58.2 & 41.8 & 74.9 & 67.4 & 48.2 & 74.2 & 61.4 & 64.3 & 45.3 & 54.8 \\
         DANN~\cite{Ganin2017Domain}         & 43.8 & 67.9 & 77.5 & 63.7 & 59.0 & 67.6 & 56.8 & 37.1 & 76.4 & 69.2 & 44.3 & 77.5 & 61.7 & 73.8 & 51.0 & 62.4 \\
         PADA~\cite{cao2018partial}          & 52.0 & 67.0 & 78.7 & 52.2 & 53.8 & 59.0 & 52.6 & 43.2 & 78.8 & 73.7 & 56.6 & 77.1 & 62.1 & 76.5 & 53.5 & 65.0 \\
         IWAN~\cite{zhang2018importance}     & 53.9 & 54.5 & 78.1 & 61.3 & 48.0 & 63.3 & 54.2 & 52.0 & 81.3 & 76.5 & 56.8 & 82.9 & 63.6 & 71.3 & 48.6 & 60.0 \\
         SAFN~\cite{xu2019larger}            & 58.9 & 76.3 & 81.4 & 70.4 & 73.0 & 77.8 & 72.4 & 55.3 & 80.4 & 75.8 & 60.4 & 79.9 & 71.8 & - & 67.7 & -  \\
         DMP~\cite{luo2022DMP}      & 54.0 & 71.9 & 81.3 & 63.2 & 61.6 & 70.0 & 62.3 & 49.5 & 77.2 & 73.4 & 54.1 & 79.4 & 66.5 & - & 67.6 & - \\
         AGAN~\cite{AGAN_2022}         & 56.4 & 77.3 & 85.1 & 74.2 & 73.8 & 81.1 & 70.8 & 51.5 & 84.5 & 79.0 & 56.8 & 83.4 & 72.8 & 80.5 & 67.7 & 74.1 \\
         DRCN~\cite{li2021DRCN}  &54.0 & 76.4 &  83.0 & 62.1 & 64.5 & 71.0 & 70.8 & 49.8 & 80.5 & 77.5 & 59.1 & 79.9 & 69.0 & 73.2 & 58.2 &  65.7\\
         DARL~\cite{DARL2022} & 55.3 & 80.7 & 86.4 & 67.9 & 66.2 & 78.5 & 68.7 & 50.9 & 87.7 & 79.5 & 57.2 & 85.6 & 72.1 & 79.9 & 67.8 & 73.9 \\
         AR~\cite{AR_2021} & \textbf{67.4} & 85.3 & 90.0 & 77.3 & 70.6 & 85.2 & 79.0 & \textbf{64.8} & \textbf{89.5} & 80.4 & 66.2 & 86.4 & 78.3 & 78.5 & 88.7 & 83.6 \\
         JUMBOT~\cite{JUMBOT_2021}& 62.7 & 77.5 & 84.4 & 76.0 & 73.3 & 80.5 & 74.7 & 60.8 & 85.1 & 80.2 &66.5 & 83.9 & 75.5 & - & 84.0 & -\\
         m-POT~\cite{mPOT_2022} & 64.6 & 80.6 & 87.2 & \textbf{76.4} & 77.6 & 83.6 & \textbf{77.1} & 63.7 & 87.6 & \textbf{81.4} & \textbf{68.5} & \textbf{87.4} & 78.0 & - & 87.0 & - \\
         \hline
         \textbf{SSOT} & 62.8 & \textbf{85.4} & \textbf{90.8} & 74.2 & \textbf{81.8} & \textbf{90.6} & 75.3 & 61.6 & \textbf{89.5} & 80.8 & 65.8 & 84.3 & \textbf{78.6}& \textbf{85.3} & \textbf{91.8} & \textbf{88.5} \\
         \bottomrule
    \end{tabular}
\end{table*}

\begin{table}[!tb]
    \centering
    \caption{Accuracies (\%) on Office-31 for PDA (ResNet-50).}
    \renewcommand{\tabcolsep}{0.38pc}
    \label{tab:office31-acc}
    \begin{tabular}{l|ccccccc}
         \toprule
         \textbf{Office-31} & A$\rightarrow$W& D$\rightarrow$W& W$\rightarrow$D& A$\rightarrow$D& D$\rightarrow$A& W$\rightarrow$A& Mean\\
         \hline
    Source~\cite{He2016Deep}            & 75.6 & 96.3 & 98.1 & 83.4 & 83.9 & 85.0 & 87.1 \\
    DANN~\cite{Ganin2017Domain}         & 73.6 & 96.3 & 98.7 & 81.5 & 82.8 & 86.1 & 86.5 \\
    PADA~\cite{cao2018partial}          & 86.5 & 99.3 & \textbf{100.0} & 82.2 & 92.7 & 95.4 & 92.7\\
    IWAN~\cite{zhang2018importance}     & 89.2 & 99.3 & 99.4 & 90.5 & 95.6 & 94.3 & 94.7 \\
    SAFN~\cite{xu2019larger}            & 87.5 & 96.6 & 99.4 & 89.8 & 92.6 & 92.7 & 93.1 \\
    DMP~\cite{luo2022DMP}      & 94.5 & 99.9 & \textbf{100.0} & 95.0 & 94.7 & 95.4 & 96.6\\
    AGAN~\cite{AGAN_2022}& \textbf{97.3} & \textbf{100.0} & \textbf{100.0} & 94.3 & 95.7 & 95.7 & 97.2\\
    DRCN~\cite{li2021DRCN} & 88.5 & \textbf{100.0} & \textbf{100.0} & 86.0 & 95.6 & 95.8 & 94.3 \\
    DARL~\cite{DARL2022} & 94.6 & 99.7 & \textbf{100.0} & \textbf{98.7} & 94.6  & 94.3 & 97.0 \\
    AR~\cite{AR_2021}  & 93.5 & \textbf{100.0} & 99.7 & 96.8 & 95.5 & 96.0 & 96.9\\
         \hline
         \textbf{SSOT}& \textbf{97.3} & \textbf{100.0}& \textbf{100.0} & \textbf{98.7} & \textbf{96.3} & \textbf{96.5} & \textbf{98.1}\\
         \bottomrule
    \end{tabular}
\end{table}

\begin{table}[!tb]
    \centering
    \caption{Accuracies (\%) on Image-CLEF for PDA (ResNet-50).}
    \renewcommand{\tabcolsep}{0.48pc} 
    \label{tab:imageclef-acc}
    \begin{tabular}{l|ccccccc}
         \toprule
         \textbf{ImageCLEF} & I$\rightarrow$P& P$\rightarrow$I& I$\rightarrow$C& C$\rightarrow$I& C$\rightarrow$P& P$\rightarrow$C& Mean \\
         \hline
         Source~\cite{He2016Deep}            & 78.3 & 86.9 & 91.0 & 84.3 & 72.5 & 91.5 & 84.1 \\
         DANN~\cite{Ganin2017Domain}         & 78.1 & 86.3 & 91.3 & 84.0 & 72.1 & 90.3 & 83.7 \\
         PADA~\cite{cao2018partial}          & 81.7 & 92.1 & 94.6 & 89.8 & 77.7 & 94.1 & 88.3 \\
         SAFN~\cite{xu2019larger}           & 79.5 & 90.7 & 93.0 & 90.3 & 77.8 & 94.0 & 87.5 \\
         DMP~\cite{luo2022DMP}     & 81.5 & 94.3 & 96.2 & 93.0 & 78.2 & 96.5 & 90.0 \\
         \hline
         \textbf{SSOT} & \textbf{84.2} & \textbf{96.7} & \textbf{99.0} & \textbf{97.0} & \textbf{83.5} & \textbf{98.7} & \textbf{93.2}  \\
         \bottomrule
    \end{tabular}
\end{table}

In this section, we evaluate the effectiveness of our SSOT in dealing with the PDA problem and show the comparisons between SSOT and existing methods. We also provide parameter sensitivity, ablation study, feature visualization, class weight visualization, and optimization comparison to analyze the proposed framework.

\subsection{Datasets and Implementation Details}
SSOT is evaluated on four adaptation datasets.

\textbf{Office-31} \cite{saenko2010adapting} consists of 3 domains with 31 classes, \textit{i.e.}, Amazon (A), Webcam (W), and Dslr (D). For the partial setting, the 10 common classes between Office-31 and Caltech-256~\cite{griffin2007caltech} are utilized for the target domain.

\textbf{Image-CLEF}\footnote{https://www.imageclef.org/2014/adaptation} consists of 3 domains with 12 classes, \textit{i.e.}, Caltech (C), ImageNet (I), and Pascal (P), which are collected from datasets Caltech-256~\cite{griffin2007caltech}, ImageNet ILSVRC 2012~\cite{russakovsky2015imagenet}, and Pascal VOC 2012~\cite{everingham2010pascal}. For the partial setting, the first 6 classes in alphabetical order are utilized for the target domain.

\textbf{Office-Home}~\cite{Venkateswara2017Deep} consists of 4 domains with 65 classes, \textit{i.e.}, Art (Ar), Clipart (Cl), Product (Pr), and Real World (Rw). These 15,500 images are mostly from an office or home environment. For the partial setting, the first 25 classes in alphabetical order are utilized as the target domain.

\textbf{VisDA-2017}~\cite{peng2017visda} is a large-scale challenging dataset, which consists of 280K images from two domains, \textit{i.e.}, {S} (synthetic-image) and {R} (real-image). The domains have 12 classes. We conduct tasks S$\rightarrow$R6 and R$\rightarrow$S6 for PDA. The first 6 classes in alphabetical order are utilized for the target domain.

The network backbones and basic settings are specified as follows. The feature extractor $f(\cdot)$ is obtained by replacing the fully-connected layers in ResNet-50~\cite{He2016Deep} with two or three fully-connected layers (2048→1024→512(→256)). The classifier $\eta(\cdot)$ is built upon the outputs of $f(\cdot)$, which consists of a single fully-connected layer with $K$ output units and a softmax activate function. The Kantorovich potential is parameterized with two fully-connected layers (512/256$\rightarrow$256$\rightarrow$1). The whole network of SSOT is implemented on the PyTorch platform and trained by the Adam optimizer. ResNet-50 is initialized by pre-training on ImageNet~\cite{russakovsky2015imagenet}, and LeNet is initialized by random values. The parameter $\varepsilon$ for the semi-dual OT formulation is set as 1. As for the inputs, we apply 224$\times$224 center crops of 256$\times$256 resized images on each dataset. The mini-batch size of the source and target domains are both set as 32.

\begin{figure}[!t]
\subfigure[Task P$\rightarrow$I]{
    \label{fig:param-p2i}
    \begin{minipage}[b]{.453\linewidth}
    \centering
    \includegraphics[scale=0.3]{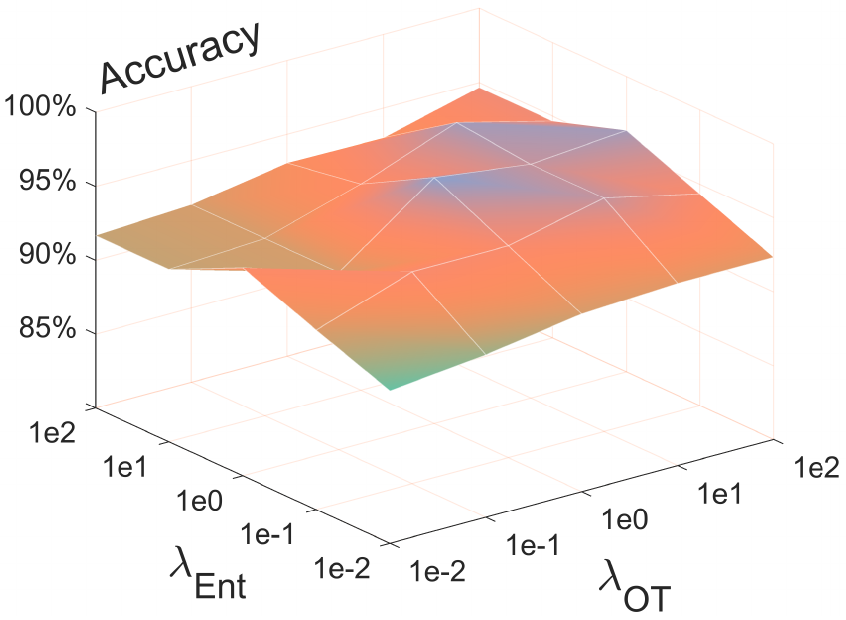}
    \end{minipage}}
\subfigure[Task A$\rightarrow$D]{\label{fig:param-A2D}
    \begin{minipage}[b]{.453\linewidth}
    \centering
    \includegraphics[scale=0.3]{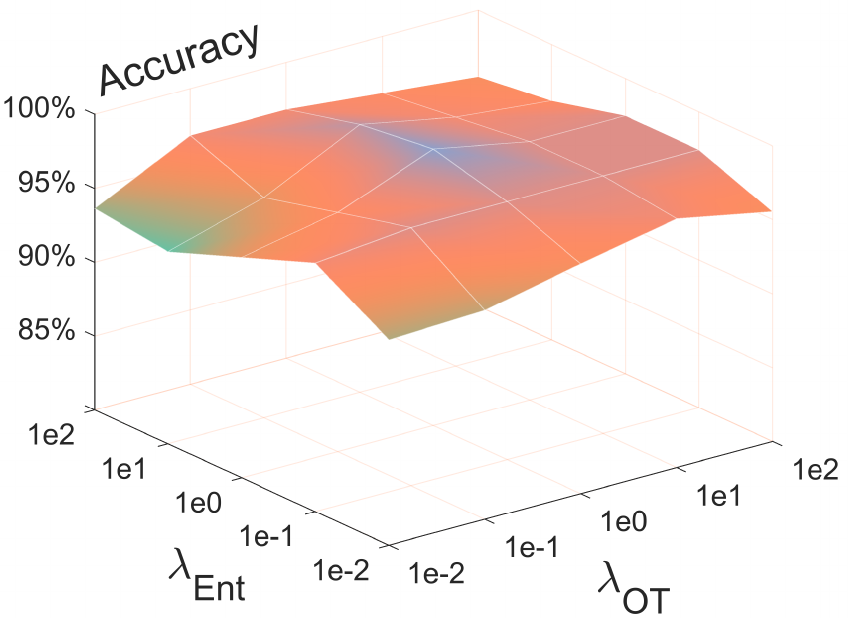}
    \end{minipage}}
\caption{Hyper-parameter sensitivity of $\lambda_{\rm Ent}$ and $\lambda_{\rm OT}$ on ImageCLEF and Office-31. Best viewed in color.}
    \label{fig:param-sensitivity}
\end{figure}

\subsection{Results and Analysis}

\textbf{Comparison.} To evaluate the model performance, we report the classification results of SSOT and make a comparison with several state-of-the-art PDA approaches. The compared methods can be roughly categorized into two groups. 

\begin{figure*}[!htb]
\subfigure[Source (P$\rightarrow$I)]{\label{fig:tsne-source-p2i}
   \begin{minipage}[b]{0.23\linewidth}
   \centering\scalebox{0.35}{ 
   \includegraphics{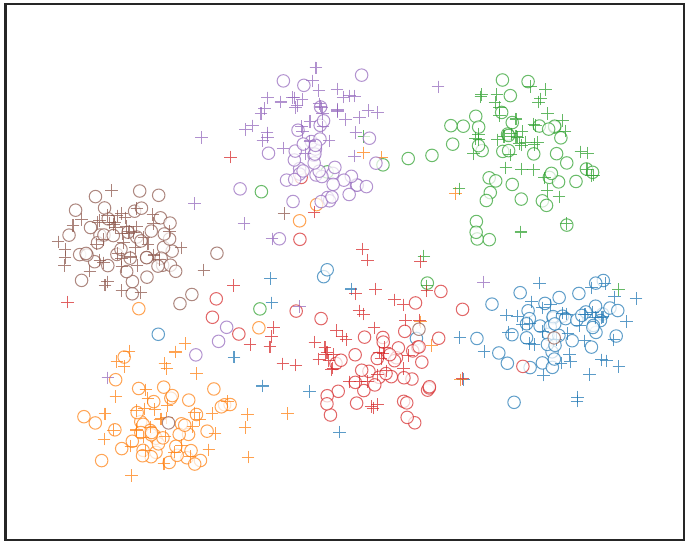}}
   \end{minipage}}
\subfigure[PADA (P$\rightarrow$I)]{\label{fig:tsne-pada-p2i}
   \begin{minipage}[b]{0.23\linewidth}
   \centering\scalebox{0.35}{
   \includegraphics{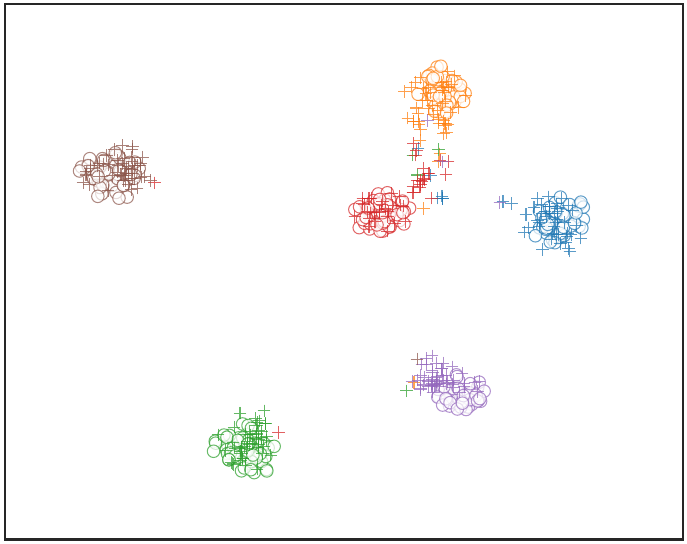}}
   \end{minipage}}
\subfigure[AR (P$\rightarrow$I)]{\label{fig:tsne-ar-p2i}
   \begin{minipage}[b]{0.23\linewidth}
   \centering\scalebox{0.335}{
   \includegraphics{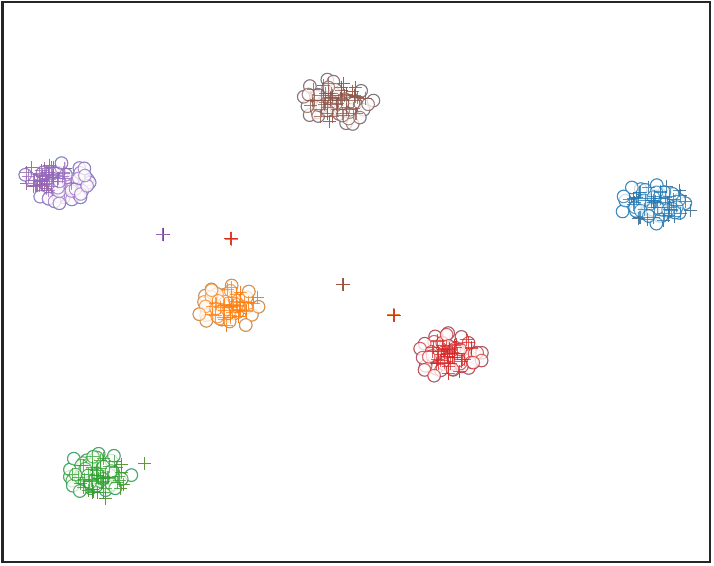}}
   \end{minipage}}
\subfigure[SSOT (P$\rightarrow$I)]{\label{fig:tsne-ssot-p2i}
   \begin{minipage}[b]{0.23\linewidth}
   \centering\scalebox{0.35}{ 
   \includegraphics{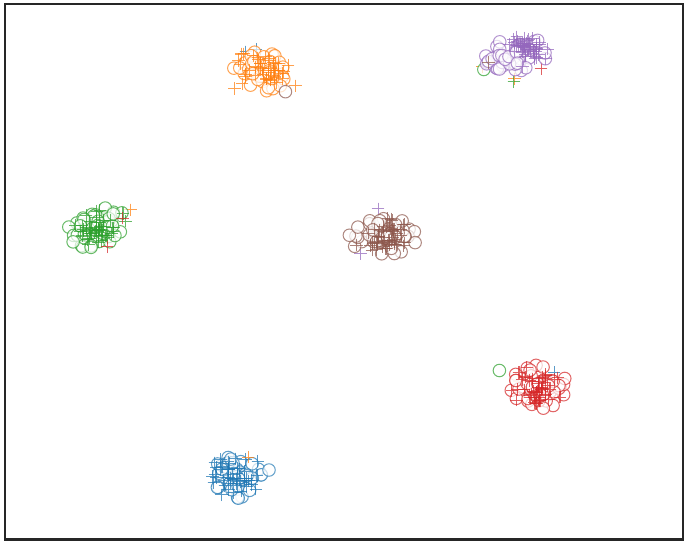}}
   \end{minipage}}
\\
\subfigure[Source (A$\rightarrow$D)]{\label{fig:tsne-source-a2d}
   \begin{minipage}[b]{0.23\linewidth}
   \centering\scalebox{0.35}{ 
   \includegraphics{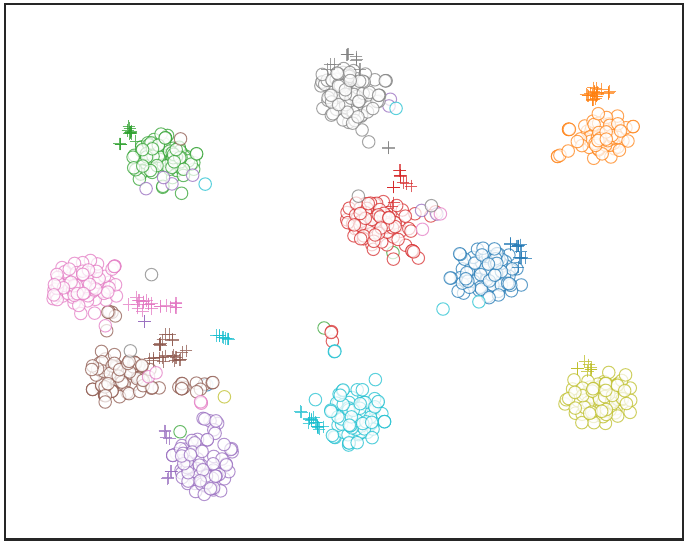}}
   \end{minipage}}
\subfigure[PADA (A$\rightarrow$D)]{\label{fig:tsne-pada-a2d}
   \begin{minipage}[b]{0.23\linewidth}
   \centering\scalebox{0.35}{
   \includegraphics{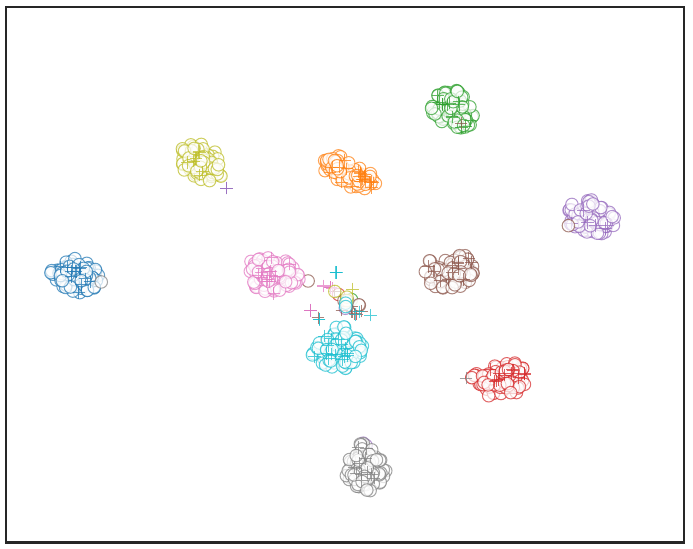}}
   \end{minipage}}
\subfigure[AR (A$\rightarrow$D)]{\label{fig:tsne-ar-a2d}
   \begin{minipage}[b]{0.23\linewidth}
   \centering\scalebox{0.335}{
   \includegraphics{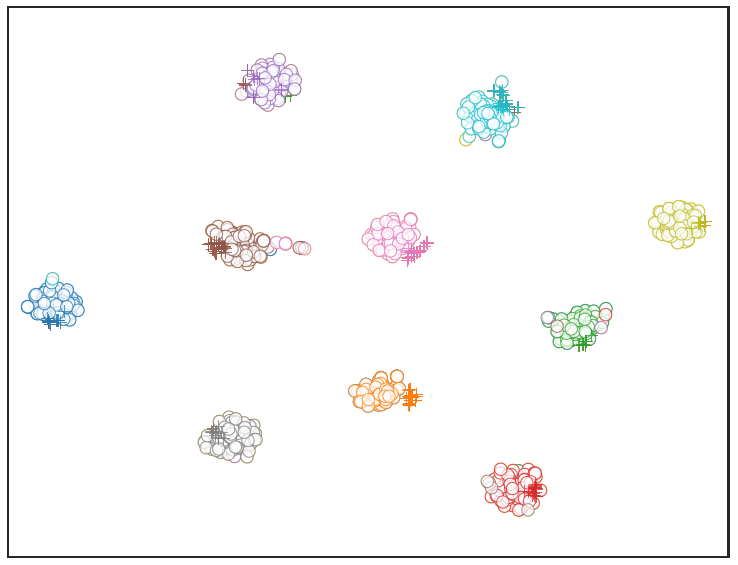}}
   \end{minipage}}
\subfigure[SSOT (A$\rightarrow$D)]{\label{fig:tsne-ssot-a2d}
   \begin{minipage}[b]{0.23\linewidth}
   \centering\scalebox{0.35}{ 
   \includegraphics{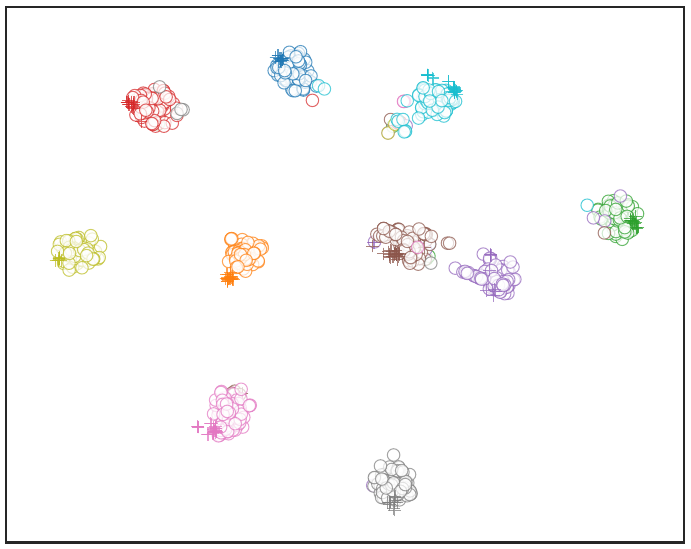}}
   \end{minipage}}
\caption{The t-SNE visualizations of features generated by Source, PADA, AR, and SSOT on Image-CLEF task P$\rightarrow$I and Office-31 task A$\rightarrow$D, respectively. Here, ``o'' means source domain, and ``+'' means target domain. Each color denotes one class. Best viewed in color.}
\label{fig:tsne-pda}
\end{figure*}

\begin{figure}[!t]
\subfigure[Task P$\rightarrow$I]{\label{fig:pda-ablation-P2I}
    \begin{minipage}[b]{0.48\linewidth}
    \centering\scalebox{0.32}{ 
    \includegraphics{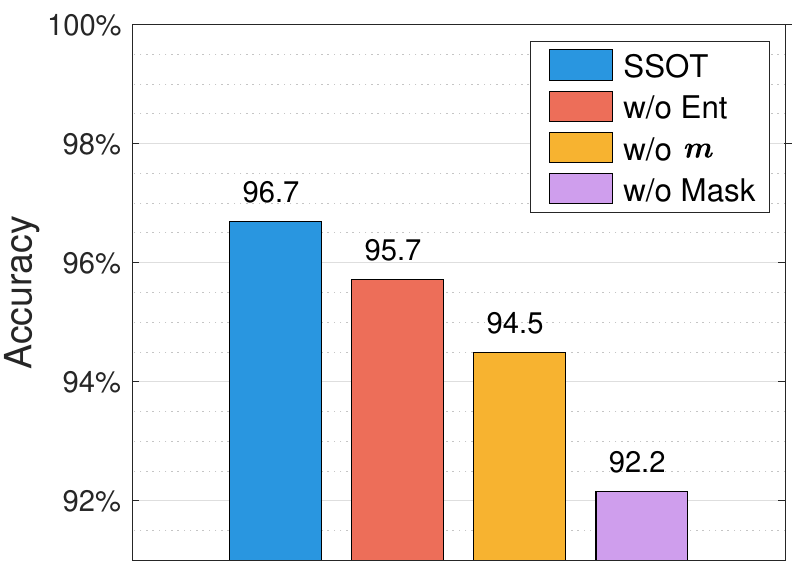}}
    \end{minipage}}
\subfigure[Task on A$\rightarrow$D]{\label{fig:pda-ablation-A2D}
    \begin{minipage}[b]{0.48\linewidth}
    \centering\scalebox{0.32}{
    \includegraphics{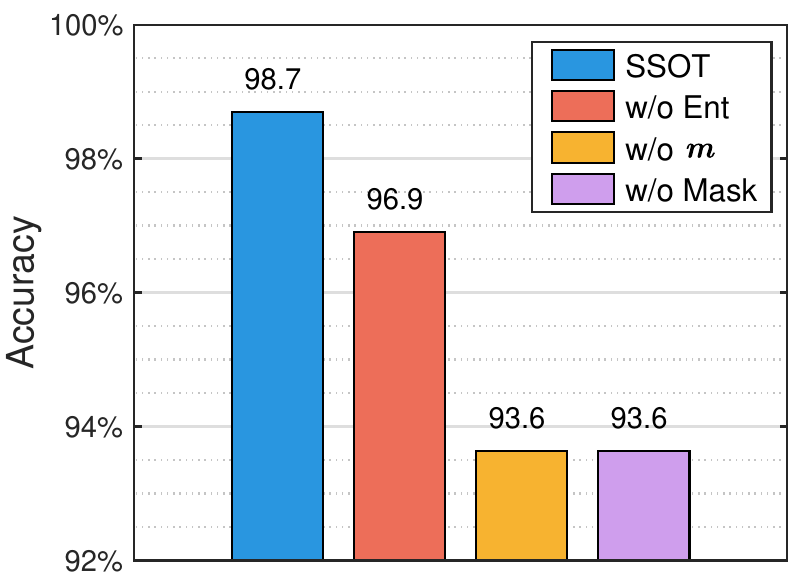}}
    \end{minipage}}
\caption{Ablation study on Image-CLEF task P$\rightarrow$I and Office-31 task A$\rightarrow$D.}
    \label{fig:pda-ablation}
\end{figure}

The results on Office-31 and ImageCLEF are presented in Tables~\ref{tab:office31-acc}~and~\ref{tab:imageclef-acc}, respectively. We notice that the adversarial method DANN performs worse than the Source model on most tasks of the two datasets. Since DANN is specifically proposed for dealing with the UDA problem, such results can indicate the existence of negative transfer, and only aligning domain marginal distributions is not enough to address the PDA problem. Then, it is reasonable for the feature norm-based method SAFN and manifold method DMP to promote the performance over the Source model since they reduce the effect of outlier classes by learning more discriminative features. Extended from DANN, PADA, and IWAN improve the performance by employing a weighting scheme to identify the outlier source samples and learn domain-invariant representations in the shared feature space. In Table~\ref{tab:office31-acc}, DARL and AGAN achieve higher average accuracies than PADA and IWAN since they explore the structure of domains and seek a class-wise domain alignment. Overall, our SSOT achieves the best performance on both datasets with average accuracies of 98.1\% and 93.2\%, respectively. Specifically, SSOT consistently outperforms other baselines on all the ImageCLEF tasks by a large margin. Besides, SSOT exceeds other baselines on all Office-31 tasks, where the accuracies on task D$\rightarrow$W and W$\rightarrow$D are both 100\%. Such results demonstrate that SSOT effectively mitigates the negative transfer and reduces the domain discrepancy.

The results on Office-Home are presented in the left of Table~\ref{tab:office-home-visda-acc}. Transfer tasks in Office-Home have a more severe negative transfer problem since there are 40 outlier classes in the source domain. PADA and IWAN gain limited improvements over DANN since they focus on domain adversarial learning while ignoring the class-wise domain alignment in the shared feature space. MMD-based method DRCN and manifold-based method DMP increase the average accuracy to 66.0\%$\sim$67.0\%, which shows that exploiting intra-domain and inter-domain structure information can encourage positive transfer. Thus, it is reasonable for the feature norm-based method SAFN and graph-based method AGAN to significantly promote the average accuracy to 71.8\% and 72.8\%, respectively.
We notice that the OT-based methods AR, JUMBOT, m-POT, and SSOT achieve significantly higher mean accuracies than the adversarial methods (\textit{e.g.}, AGAN, DRCN, and DARL), which shows the superiority of OT distance in characterizing the domain discrepancy of the PDA problem. Unlike other OT-based methods, SSOT further characterizes the class-wise structure of domains via a masked OT distance. We can observe that SSOT surpasses other baselines with a mean accuracy of 78.6\%. Such results indicate that SSOT is helpful in reducing negative transfer via the masked OT distance.

The results on VisDA-2017 are presented in the right of Table~\ref{tab:office-home-visda-acc}. Considerable domain gaps between synthetic and real samples are explored. In this situation, the mean accuracies of AGAN and DARL are much higher than PADA, IWAN, and DRCN, which also indicates that exploring the intrinsic structure of domains is crucial for positive transfer. Besides, OT-based methods AR, JUMBOT, m-POT, and SSOT provide comparable improvements over other baselines, which further confirms the superiority of the OT metric. SSOT outperforms other baselines notably and increases the average accuracy to 88.5\%. The accuracy of SSOT is higher than the second-best method AR by 3.1\% on task S$\rightarrow$R6. Overall, we can conclude that SSOT is effective in reducing the domain discrepancy on challenging datasets.

\textbf{Parameter Sensitivity.} We investigate the selection of hyper-parameters $\lambda_{\rm Ent}$ and $\lambda_{\rm OT}$ on ImageCLEF task P$\rightarrow$I and Office-31 task A$\rightarrow$D. The two parameters are used to balance the target entropy loss $\mathcal{L}_{\rm Ent}$ and the OT-based domain adaptation loss $\mathcal{L}_{\rm OT}$. We search the parameters from the pre-defined set $\{$1$e$-3,1$e$-2,1$e$-1,1$e$0,1$e$1$\}$. The grid search results are shown in Fig.~\ref{fig:param-sensitivity}. We can observe that the peak areas, \textit{i.e.}, highest accuracies, can be achieved with $\lambda_{\rm Ent}$=$\{$1$e$0,1$e$1$\}$, $\lambda_{\rm OT}$=$\{$1$e$0,1$e$1$\}$. Additionally, the accuracies around the peak regions will decrease with smaller values of $\lambda_{\rm OT}$, which demonstrates that the OT-based domain adaptation loss is indeed necessary for achieving better performance.

\textbf{Ablation Study.} We evaluate the effectiveness of different modules in SSOT and show the results in Fig.~\ref{fig:pda-ablation}. SSOT without importance weights $\boldsymbol{m}$, mask mechanism and target entropy loss are abbreviated as ``w/o $\boldsymbol{m}$'', ``w/o Mask'' and ``w/o Ent'', respectively. From the results, we observe that the full model SSOT achieves the best performance, which indicates that each module is helpful in the adaptation process. Besides, the accuracies of SSOT w/o Ent are higher than SSOT w/o $\boldsymbol m$ and SSOT w/o Mask, which demonstrates that the importance weights and mask mechanisms play more important roles in SSOT. Besides, SSOT w/o $\boldsymbol m$ gives comparable results with w/o Mask, which demonstrates that the mask mechanism is also effective in decreasing the negative transfer of PDA.

\textbf{Feature Visualization.} To provide an intuitive understanding of the aligned features, we use t-distribution Stochastic Neighbour Embedding (t-SNE)~\cite{van2008tSNE} to visualize the features generated by the Source, PADA, AR, and SSOT methods.

As shown in Fig.~\ref{fig:tsne-pda}, we conduct such experiments on Image-CLEF and Office-31. To observe the misalignment across classes, source features belonging to the shared classes are selected, and the visualizations are colored at class-level. In Fig.~\ref{fig:tsne-source-p2i}-\subref{fig:tsne-source-a2d}, we can observe that the spatial distributions of different domains remain different, and there is no clear decision boundary between classes. In Fig.~\ref{fig:tsne-pada-p2i}-\subref{fig:tsne-ssot-p2i}, we can observe that PADA, AR, and SSOR all can improve the results of the Source model and seek a domain alignment in the shred feature space. However, there is a local confusion among the red, blue, and orange clusters in Fig.~\ref{fig:tsne-pada-p2i}. In Fig.~\ref{fig:tsne-ar-p2i}, some target samples are also falling outside the cluster, far away from their corresponding class centers.
In comparison, our SSOT has better intra-class inter-inter-class separability and intra-class compactness, as shown in Fig.~\ref{fig:tsne-ssot-p2i}. For PADA on Office-31 task A$\rightarrow$D, some target samples are indistinguishable, as shown in the middle of Fig.~\ref{fig:tsne-pada-a2d}. The reason may be that PADA exploits adversarial learning to obtain domain-invariant features but ignores the discriminative structure of domains. Compared with PADA, AR separates the clusters with larger margins in Fig.~\ref{fig:tsne-ar-a2d} since it learns the class weights of the source domain by minimizing the Wasserstein distance. Comparatively, our SSOT further explores the discriminative structure of the latent feature space by incorporating label information into the OT distance. In Fig.~\ref{fig:tsne-ssot-a2d}, scatters from the same class but different domains are closer than PADA. The results indicate that SSOT, which reweighs the OT distance and mitigates the label shift via label information, obtains more transferable and discriminative features under the partial setting.

\begin{figure}[!htb]
\label{fig:weight-visualization-pda}
\centering
\subfigure[Weights Comparison on task P$\rightarrow$I]{\label{fig:class-weight-p2i}
    \begin{minipage}[b]{0.98\linewidth}
    \scalebox{0.32}{
    \includegraphics{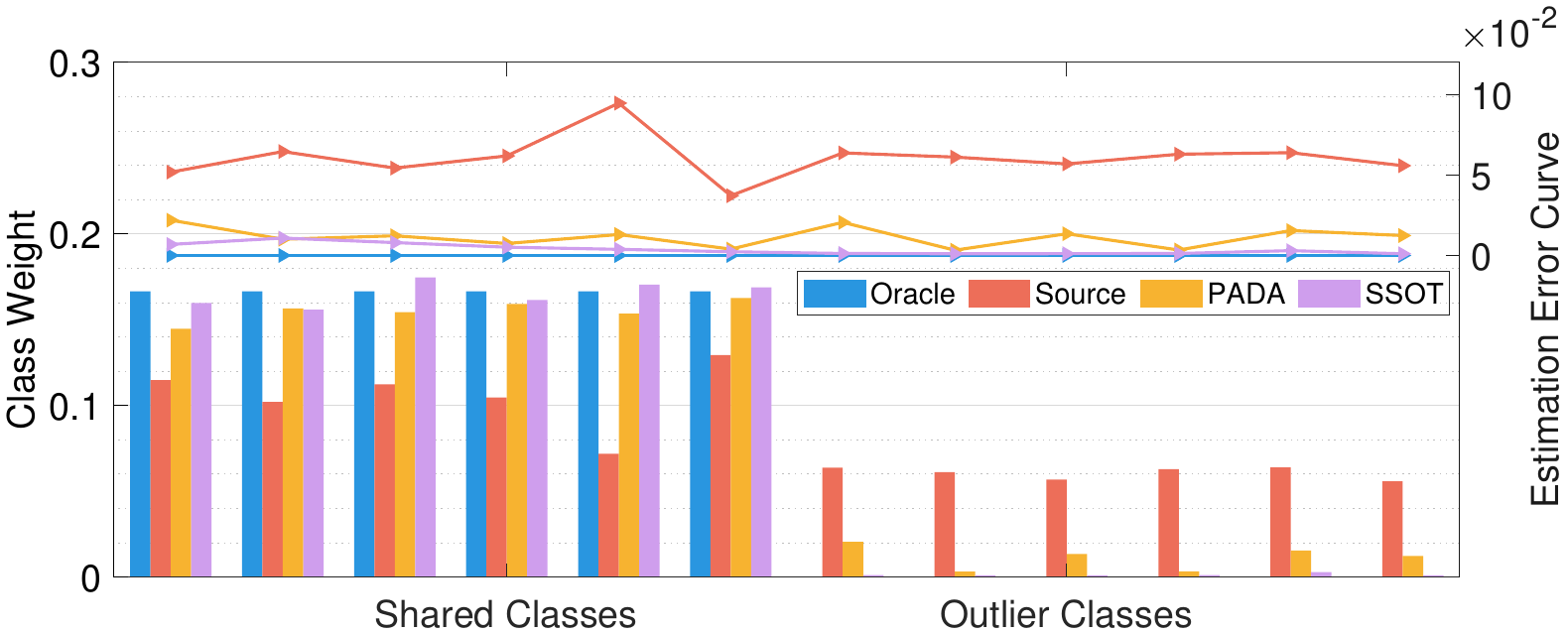}}
    \end{minipage}}\\
\subfigure[Weights Comparison on task A$\rightarrow$D]{\label{fig:class-weight-a2d}
    \begin{minipage}[b]{0.98\linewidth}
    \scalebox{0.32}{
    \includegraphics{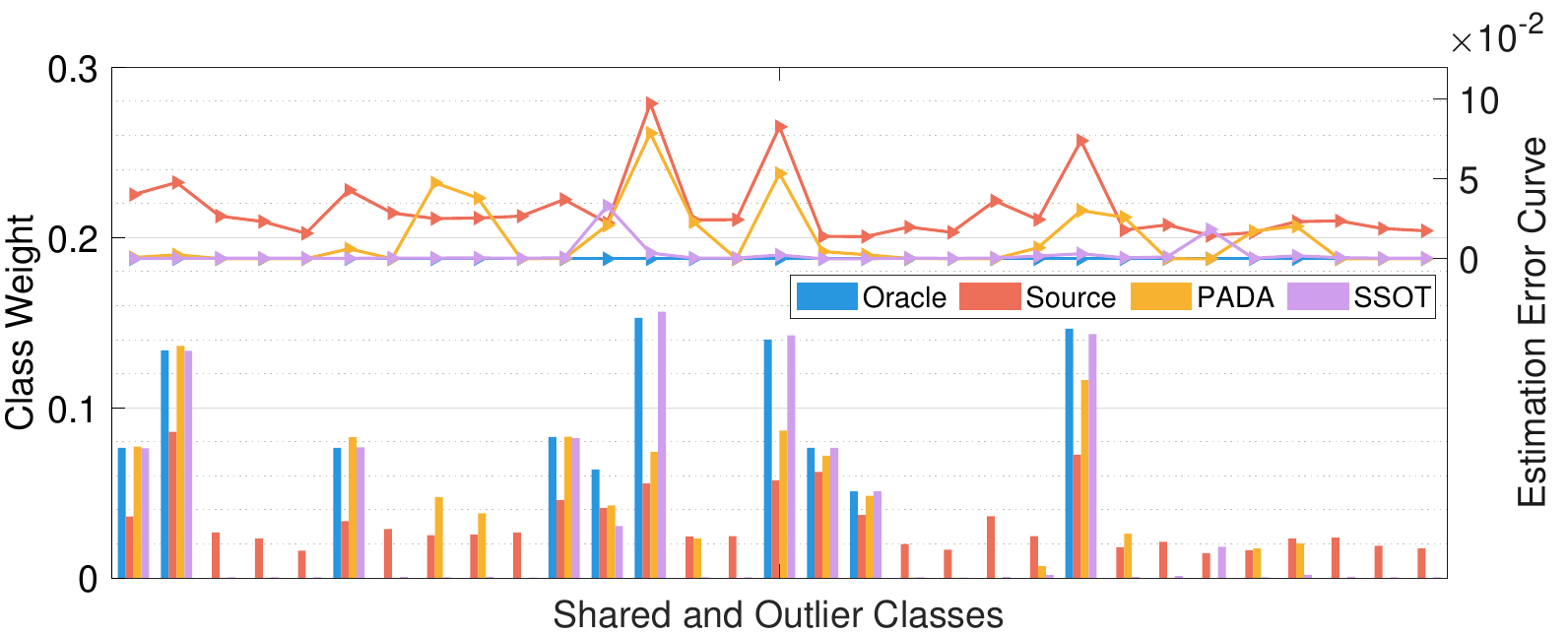}}
    \end{minipage}}
\caption{Class weights visualization and estimation error curve of Oracle, Source, PADA, and SSOT on Image-CLEF task P$\rightarrow$I and Office-31 task A$\rightarrow$D, respectively. Best viewed in color.}
\end{figure}

\textbf{Class Weight Visualization.} To provide an intuitive observation of the class weights estimated by different methods, we compare the estimation $\hat{\boldsymbol{p}}^T_Y$  with the ``Oracle'' (\textit{i.e.}, ground truth ${\boldsymbol p}_{Y}^T$) class weights of the target domain. Additionally, the estimation error curve is computed by $\left|\hat{\boldsymbol p}_{Y}^T - {\boldsymbol p}_{Y}^T \right|$. The ``Oracle'' estimation error is 0.

Fig.~\ref{fig:class-weight-p2i}-\ref{fig:class-weight-a2d} show comparisons on tasks P$\rightarrow$I (Image-CLEF) and A$\rightarrow$D (Office-31), where ``Shared'' represents these common classes across domains and ``Outlier'' represents these source-only classes. For Image-CLEF, the class weight is a uniform distribution of the shared classes. We notice that the Source model, PADA, and SSOT all provide higher weights on the shared classes while lower weights on the outlier classes. Specifically, the class weights of SSOT on the outlier classes are too small (about $1e$-3) to be visible. Corresponding error curves of SSOT are nearly zero, which also indicates that the estimated class weights of SSOT are more similar to the Oracle. Office-31 has a severe label shift problem due to non-uniform class weights and more outlier classes. The large difference between the histogram of the Source model and the Oracle one indicates that it is necessary to identify and filter out these outlier classes. The class weights of SSOT are most similar to the Oracle's on different classes. Besides, the class weights of SSOT on the outlier classes are also too small (about $1e$-6 $\sim1e$-4) to be visible. Compared with the Source model and PADA, the error curve of SSOT is closer to the Oracle one, which further validates that SSOT can deal with the label shift problem better. A more accurate class weights estimation will decrease the effect of negative transfer and enhance the discriminative structure of the shared classes.

\begin{figure}[!htb]
\begin{minipage}[b]{0.98\linewidth}
\centering\scalebox{0.36}{
\includegraphics{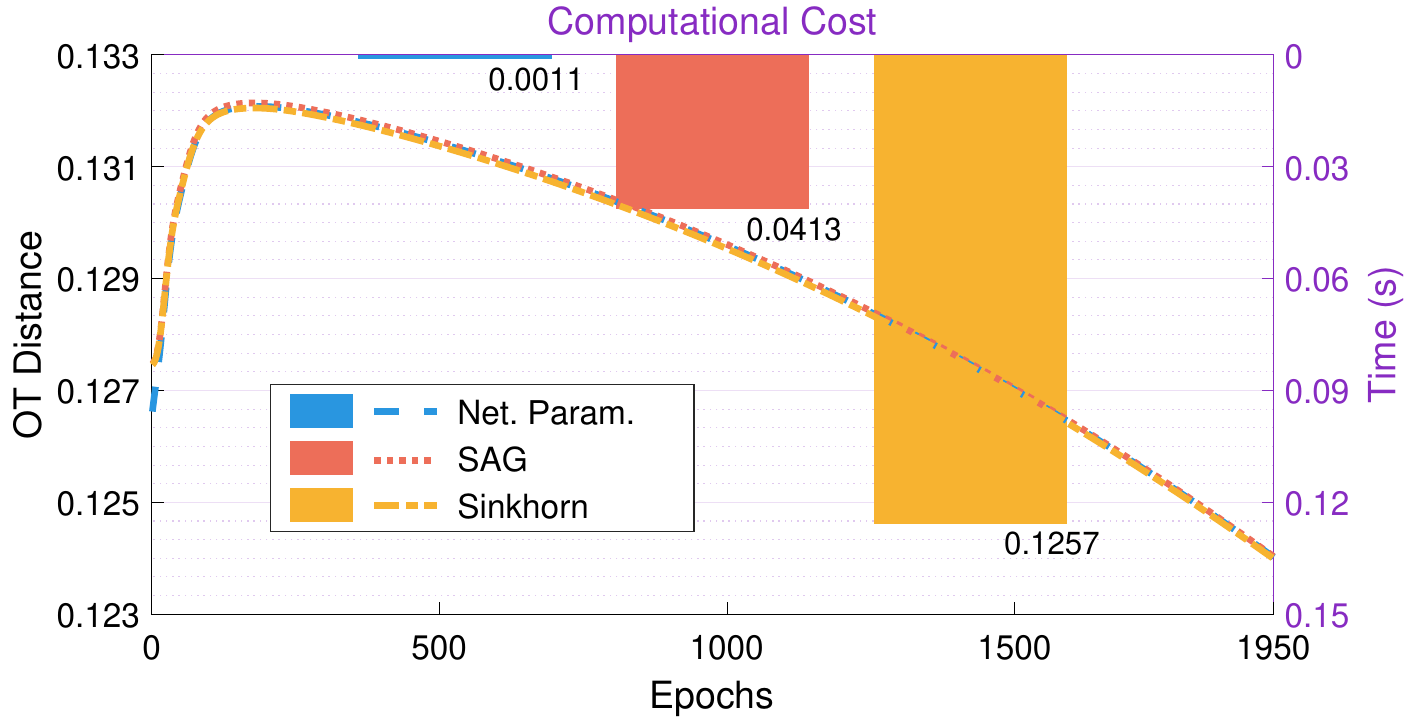}}
\end{minipage}
\caption{OT-solver comparison on Office-31 task A$\rightarrow$D. The OT distance and computational time (s) are obtained from Network Parameterization (Net. Param.), SAG, and Sinkhorn algorithms for SSOT. Best viewed in color.}
\label{fig:w-distance}
\end{figure}

\textbf{Optimization Comparison.} To verify the effectiveness of network parameterization of the OT solver, we compared it with stochastic averaged gradient (SAG)~\cite{genevay2016stochastic} and Sinkhorn~\cite{cuturi2013sinkhorn} algorithms on
Office-31 task A$\rightarrow$D. Specifically, we compare the OT distance curves w.r.t different epochs and computational time of each algorithm. All experiments are run on a device with an NVIDIA GTX1080Ti GPU. In Fig.~\ref{fig:w-distance}, we find that the three algorithms have consistent OT distance curves, which proves that the network-based OT solver can also approximate the OT distance. The OT distance calculates the discrepancy between the reweighed source domain and the target domain. Due to the influence of source outlier classes, the OT distance is increasing in the beginning. By learning domain-invariant and discriminative features in the shared feature space, we can notice that the OT distance is getting smaller and smaller. Although the curves are consistent, the network parameterization algorithm takes the shortest time with 0.0011s per epoch. These results prove that our network-based OT solver in SSOT is more efficient.

\section{Conclusion}\label{sect:conclusion}

In this paper, we consider the significant label shift in PDA and propose an OT-based method called Soft-masked Semi-dual Optimal Transport (SSOT) to solve the problem. To identify the source outlier classes and mitigate the label shift across domains, we incorporate an importance weighting scheme and provide a reweighed source domain. Besides, we construct a soft mask matrix to reweigh the elements in the cost matrix, which can promote positive transportation between intra-class samples and achieve a class-wise domain alignment in the shared feature space. To deal with large-scale OT problems, a semi-dual OT formulation is employed to reduce the domain discrepancy between the reweighed source domain and the target domain. Further, the dual variable is parameterized with the Kantorovich network, which allows an efficient and accurate approximation of the OT solution. Extensive experiment results validate the effectiveness of SSOT.

An interesting future direction is to explore a reweighed unbalanced optimal transport algorithm for PDA, which may be more robust with label shift across domains.

\section*{References}

\bibliography{ref-ssot}
\bibliographystyle{IEEEtran}
\end{document}